\documentclass{article}
\pdfoutput=1
\usepackage{microtype}
\usepackage{graphicx}
\usepackage{subcaption}
\usepackage{booktabs} % for professional tables

\usepackage{hyperref}

\usepackage[accepted]{icml2024}

\usepackage{amsmath}
\usepackage{amssymb}
\usepackage{mathtools}
\usepackage[normalem]{ulem}
\usepackage{float}

\usepackage[capitalize,noabbrev]{cleveref}

\newcommand{\revbz}[1]{\textcolor{black}{#1}}

\usepackage[textsize=tiny]{todonotes}

\usepackage{siunitx}

\newcommand{\prepar}{\vspace{-0.05in}}
\newcommand{\trueM}[0]{M^\dagger}

\newcommand{\estM}[0]{\hat{M}}

\newcommand{\Tinterval}{\mathcal{T}}
\newcommand{\Tcontext}{\mathcal{T}_\mathrm{context}}
\newcommand{\Tpred}{\mathcal{T}_\mathrm{pred}}
\newcommand{\estY}{\hat{Y}}
\newcommand{\estQ}{\hat{Q}}
\newcommand{\pred}{\mathrm{pred}}
\newcommand{\context}{\mathrm{context}}
\newcommand{\causal}{\mathrm{causal}}
\newcommand{\hybrid}{\mathrm{hybrid}}
\newcommand{\bestI}{\mathcal{I}^*}
\newcommand{\estbestI}{\hat{\mathcal{I}}}

\usepackage{xspace}
\newcommand{\modelname}{H$^2$NCM\xspace}

\icmltitlerunning{Hybrid Squared Neural ODE Causal Modeling}

\begin{document}

\twocolumn[
\icmltitle{Hybrid$^2$ Neural ODE Causal Modeling and an Application to Glycemic Response}

%\mel{I think "Square" obscures/makes confusing what is done in this work (I picture square impulses or something). I'm no branding expert though.}

%\mel{Another high-level concern---The loss functions are currently not well-defined because solutions to the ODEs are not well-defined...because we have not specified initial conditions. @Bob: what did you do about this? Can you try to write this up mathematically and we can incorporate? UPDATE: Bob said he trained an LSTM to estimate i.c.'s from historical data, and this is in Sec. 5. We need to develop notation for IC's; put them in our equations; and use them when specifying a "solution" to an ODE.}



\icmlsetsymbol{equal}{*}

\begin{icmlauthorlist}
\icmlauthor{Bob Junyi Zou}{icme}
\icmlauthor{Matthew E. Levine}{ewsc}
\icmlauthor{Dessi P. Zaharieva}{ped}
\icmlauthor{Ramesh Johari}{msae}
\icmlauthor{Emily B. Fox}{csstat,czbio}
\end{icmlauthorlist}

\icmlaffiliation{icme}{Institute for Computational and Mathematical Engineering, Stanford University}%, Stanford, CA 94305,USA}
\icmlaffiliation{ewsc}{%Eric and Wendy Schmidt Center, 
Broad Institute of MIT and Harvard} %, Cambridge, MA 02142, USA}
\icmlaffiliation{ped}{Department of Pediatrics, Stanford University}%, Stanford, CA 94304, USA}
\icmlaffiliation{msae}{Department of Management Science and Engineering, Stanford University}%, Stanford, CA 94305, USA}
\icmlaffiliation{csstat}{Department of Statistics and Department of Computer Science, Stanford University}%, Stanford, CA 94305, USA}
\icmlaffiliation{czbio}{Chan Zuckerberg Biohub – San Francisco}%, San Francisco, USA}

\icmlcorrespondingauthor{Emily B. Fox}{ebfox@stanford.edu}
%\icmlcorrespondingauthor{Ramesh Johari}{rjohari@stanford.edu}

\icmlkeywords{Machine Learning, ICML}

\vskip 0.3in
]



\printAffiliationsAndNotice{}  % leave blank if no need to mention equal contribution

\begin{abstract}
Hybrid models composing mechanistic ODE-based dynamics with flexible and expressive neural network components have grown rapidly in popularity, especially in scientific domains where such ODE-based modeling offers important interpretability and validated causal grounding (e.g., for counterfactual reasoning). The incorporation of mechanistic models also provides inductive bias in standard blackbox modeling approaches, critical when learning from small datasets or partially observed, complex systems.  Unfortunately, as the hybrid models become more flexible, the causal grounding provided by the mechanistic model can quickly be lost.  We address this problem by leveraging another common source of domain knowledge: \emph{ranking} of treatment effects for a set of interventions, even if the precise treatment effect is unknown.  We encode this information in a \emph{causal loss} that we combine with the standard predictive loss to arrive at a \emph{hybrid loss} that biases our learning towards causally valid hybrid models.  We demonstrate our ability to achieve a win-win, state-of-the-art predictive performance \emph{and} causal validity, in the challenging task of modeling glucose dynamics post-exercise in individuals with type 1 diabetes.
\end{abstract}

\section{Introduction}
\label{sec:intro}

In many scientific and clinical domains, an influx of high resolution sensing data has brought the promise of more refined and informed scientific discovery and decision-making. The motivating example we consider in this paper is type 1 diabetes (T1D) management, where continuous glucose monitoring (CGM) and smart insulin pumps are revolutionizing care by offering near-real-time insights into physiological states \citep{tauschmann2022ispad}. To provide data-driven management strategies, there is a pressing need for interpretable models that not only make \textbf{accurate predictions}, but also \textbf{grasp the causal mechanisms} behind physiological responses (e.g., for counterfactual reasoning over various potential interventions). The target of causally-grounded, performant models is critical in many scientific disciplines ranging from astronomy to cell biology to neuroscience; in many of these settings, we are faced with only partial or indirect observations of a complex physiological or physical process we aim to reason about.

Blackbox sequence models have demonstrated extraordinary performance in numerous fields, ranging from natural language processing to forecasting. Popular methods include recurrent neural networks (RNNs), such as long short-term memory networks (LSTMs)~\citep{hochreiter1997long} and gated recurrent units (GRUs)~\citep{chung2014empirical}; (temporal) convolutional neural networks (CNNs)~\citep{lea2016temporal,bai2018empirical,shi2023sequence}; Transformer models, including Informer~\citep{zhou2021informer} and Autoformer~\citep{wu2021autoformer}; and state space sequence models, such as S4~\citep{gu2021efficiently}. % and its diagonal approximation S4D~\citep{gu2022parameterization}. 
Despite their transformative role in many fields, the application of these models to various scientific domains has encountered challenges. An obvious hurdle is the limited size of some scientific datasets. But even in the presence of lots of data, these methods are still crippled by the incomplete nature of what can be measured. This problem is exacerbated by the fact that most data is {\em observational}: blackbox models are adept at identifying associations leading to good predictions, but may not learn causally coherent models. For example, in our T1D setting, insulin delivery frequently coincides with a planned meal or snack, which initially causes glucose to rise; the model may incorrectly infer that insulin causes glucose to rise, when the reality is the opposite.
%; the model may incorrectly learn that when an individual delivers insulin, glucose rises because the consumed carbohydrates will lower glucose. The resulting predictions may be accurate, but the causal explanation is incorrect.}%\newebf{We see exactly this in Fig.~\ref{fig:}.}

%On the other hand, mechanistic models specified via a set of ordinary differential equations (ODEs)---which encode known/validated causal relationships---are used in many scientific disciplines, including cardiac modeling \cite{davies2016recent}, drug metabolism \cite{johnson2010semi}, immunology \cite{brown2018applications},  epidemiology \cite{holmdahl2020wrong} and neural circuit \cite{ladenbauer2019inferring}.  In addition to 

Sophisticated mechanistic models---specified via a set of ordinary differential equations (ODEs)---remain a preferred method in many scientific domains, as they capture lab-validated or otherwise known physical or physiological causal properties of the system. Examples include cardiac and renal modeling~\citep{hilgemann1987excitation,ten2004model,marsh2005nonlinear,passini2017human}, immunology and viral kinetics~\citep{perelson1996hiv,canini2014viral},  epidemiology~\citep{he2020seir}, neural circuits~\citep{hodgkin1952quantitative,ladenbauer2019inferring}, and pharmacokinetics~\citep{holz2001compartment}. %\bz{Revised references: examples include cardiac modeling~\cite{hilgemann1987excitation,ten2004model}, immunology and viral kinetics~\cite{perelson1996hiv,canini2014viral,passini2017human},  epidemiology~\cite{he2020seir}, neural circuit~\cite{hodgkin1952quantitative,ladenbauer2019inferring}, compartment modeling in pharmacokinetics~\cite{holz2001compartment} and renal blood flow regulation~\cite{marsh2005nonlinear}}.
The UVA/Padova simulator~\citep{man2014uva} of insulin-glucose dynamics we consider in this paper has been FDA approved as a substitute for pre-clinical trials in the development of artificial pancreas algorithms.  %Of note, however, is that sophisticated mechanistic models tend to add states to capture more intricate, non-Markovian dynamics such as delays, while maintaining the tractability of linear ODEs, leading to over-parameterization, and as a result, nonidentifiability. 
Notably, sophisticated mechanistic models often introduce additional states to capture intricate, non-Markovian dynamics such as delays. While this approach preserves the tractability of linear ODEs, it can lead to over-parameterization. % and, consequently, non-identifiability.}
%\delrj{over-parameterization, over-specification, \newmel{and practical unidentifiabilities}}. 
The latest version of the UVA/Padova simulator has over 30 states.  Yet, %While highly parameterized, 
these models fail to capture important dynamics observed in real-world data; moreover, their parameter inference is highly sensitive.
%\delrj{these models are brittle and tend to struggle to model noisy, real-world data collected under conditions that do not match the controlled laboratory or simulated environment settings.} 
Further limiting this class of models is the typical assumption of a fixed mechanistic structure with a static set of simulator parameters.  In T1D, many unobserved or partially-observed time-varying factors affect glycemic responses, including stress, hormone cycles, sleep, and activity levels~\citep[cf.,][]{wellen2005inflammation}. Likewise, the causal mechanism governing glycemic responses can vary---e.g., during certain types of physical activity, multiple underlying processes (not modeled in UVA/Padova) are activated based on available energy sources~\citep{mcardle2006essentials}.

Given these limitations, we focus on an alternative approach that hybridizes machine learning (ML) with domain knowledge encoded in mechanistic models.
These \emph{hybrid models} have gained traction across the natural and physical sciences~\citep{willard2022integrating},
while being given different names and interpretations such as \emph{graybox modeling}~\citep{rico1994continuous}, \emph{physics-informed machine learning}~\citep{karniadakis2021physics}, \emph{universal differential equations}~\citep{rackauckas2020universal} and \emph{neural closure learning}~\citep{gupta2021neural}. The core idea of hybrid modeling is to infuse domain inductive bias such that the trained model can gain the best of both worlds---mechanistic rigor with the flexibility and expressivity of deep learning. The hope is that hybrid methods not only allow one to solve complex modeling problems with improved precision and accuracy \citep{pathak2018hybrid,willard2022integrating}, but also reduce the demand on data \citep{rackauckas2020universal}, enhance reliability and robustness \citep{didona2015enhancing}, and make the ML algorithms interpretable \citep{karniadakis2021physics,du2019techniques}. Examples of successes of hybrid modeling in healthcare can be found in \citet{qian2021integrating,sottile2021real,hussain2021neural}.

\begin{figure}[t!]
    %\vskip 0.2in
    \begin{center}
    \includegraphics[width=\columnwidth]
{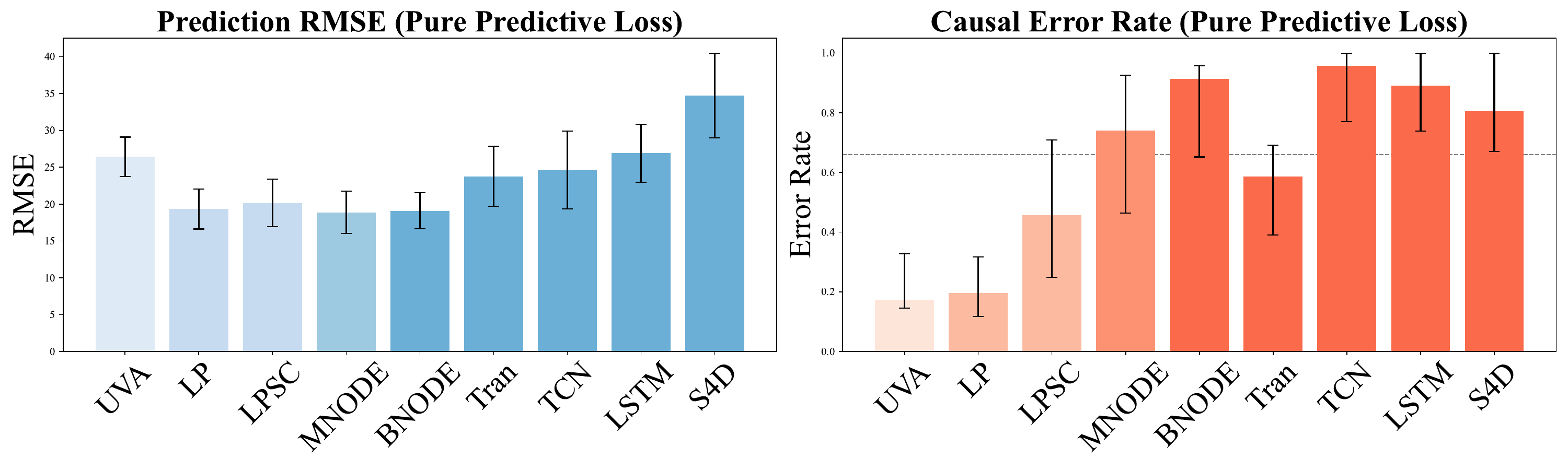}
%{sections/img/alpha0_pred_and_causal.png}
        \vskip -0.11in
    \caption{\revbz{Prediction RMSE and standard error (\emph{left}) vs. 10/50/90-th (lower/bar/upper) percentile} classification error rate (\emph{right}) for (1) UVA/Padova mechanistic model (UVA); (2) hybrid models: latent parameter dynamics (LP), latent parameter dynamics + state closure (LPSC), MNODE; and (3) blackbox models: neural ODE (BNODE), temporal convolutional network (TCN), \revbz{LSTM, Transformer (Trans)} and S4D, all trained on pure predictive loss ($\alpha=0$).  The dashed gray line in the second figure corresponds to the causal classification error rate of random guessing \revbz{($2/3$)}.}
    \label{fig:predictive-loss-results}
    \end{center}
    \vspace{-0.21in}
\end{figure}

Indeed, in Fig.~\ref{fig:predictive-loss-results} (left), we see the important inductive bias such mechanistic models provide.  On the one hand, the pure mechanistic model makes poor predictions due to its brittle nature while on the other hand, the flexible blackbox models are subject to overfitting.  The hybrid models generally outperform either alternative in terms of prediction error.

Unfortunately, as we also see in Fig.~\ref{fig:predictive-loss-results} (right), hybrid models can quickly lose their valid causal grounding as more and more flexible modeling components are deployed. When tasked with selecting the intervention with the largest treatment effect amongst a set of three counterfactual simulations (see Sec.~\ref{sec:experiments}), classification error for hybrid models gradually deteriorates as the model becomes more blackbox.  Amongst the hybrid and blackbox models, we see that the majority have a classification error rate statistically indistinguishable from random guessing, 67\%. %above $67\%$, the error rate of random guessing.}

% To address these challenges, we leverage another common source of domain knowledge:  the direction and relative magnitudes of treatment effects.  For example, while the precise glycemic response to eating a cake versus a salad may vary between and within individuals, the general finding that eating cake increases blood glucose more than salad holds consistently.  That is, we can encode additional domain knowledge through 

To address these challenges, we leverage another common source of domain knowledge: rankings of various treatment effects. While domain knowledge is often insufficient to specify expected treatment effects under counterfactually-applied interventions, we are still often able to make quantitative claims about relative magnitudes of different treatment effects. For example, while we may not know the precise treatment effect of eating a small salad versus a whole birthday cake (all else held constant), we do know the sign and relative scale of treatment effect. %\delrj{To capture this, we define expected rank orderings across a set of pre-defined interventions.} 
As this prior knowledge is challenging to encode in the hybrid model itself, we introduce a \emph{causal loss} function that encourages our learned model to perform well in these comparison tasks.  Our \emph{hybrid loss} is a convex combination of predictive loss and causal loss.  When applied to training hybrid models, we refer to the overall method as \textbf{hybrid$^2$ neural ODE causal modeling (\modelname)}.  In a real-world task of guiding individuals with T1D on the impact of interventions so they can safely exercise, we show that our \modelname approach achieves the best of both worlds: state-of-the-art predictive accuracy with causal validity.

%\textbf{To bridge this gap, in this paper, we introduce our Hybrid Square Neural Causal Learning (H2NCM) algorithm that allows \mel{balances instead of allows?} one to get models that are both causally valid and predictively accurate, and empirically demonstrate this on real-world data.} H2NCM comprises two integral components: a tailored hybrid model that embed established causal structures and a hybrid loss that reinforces causal constraints beyond the architecture's capacity in a soft and adjustable way. 

\section{The Hybrid Modeling Spectrum}
\begin{figure}[t!]
 %   \vskip 0.2in
    \begin{center}
    \centerline{\includegraphics[width=\columnwidth]{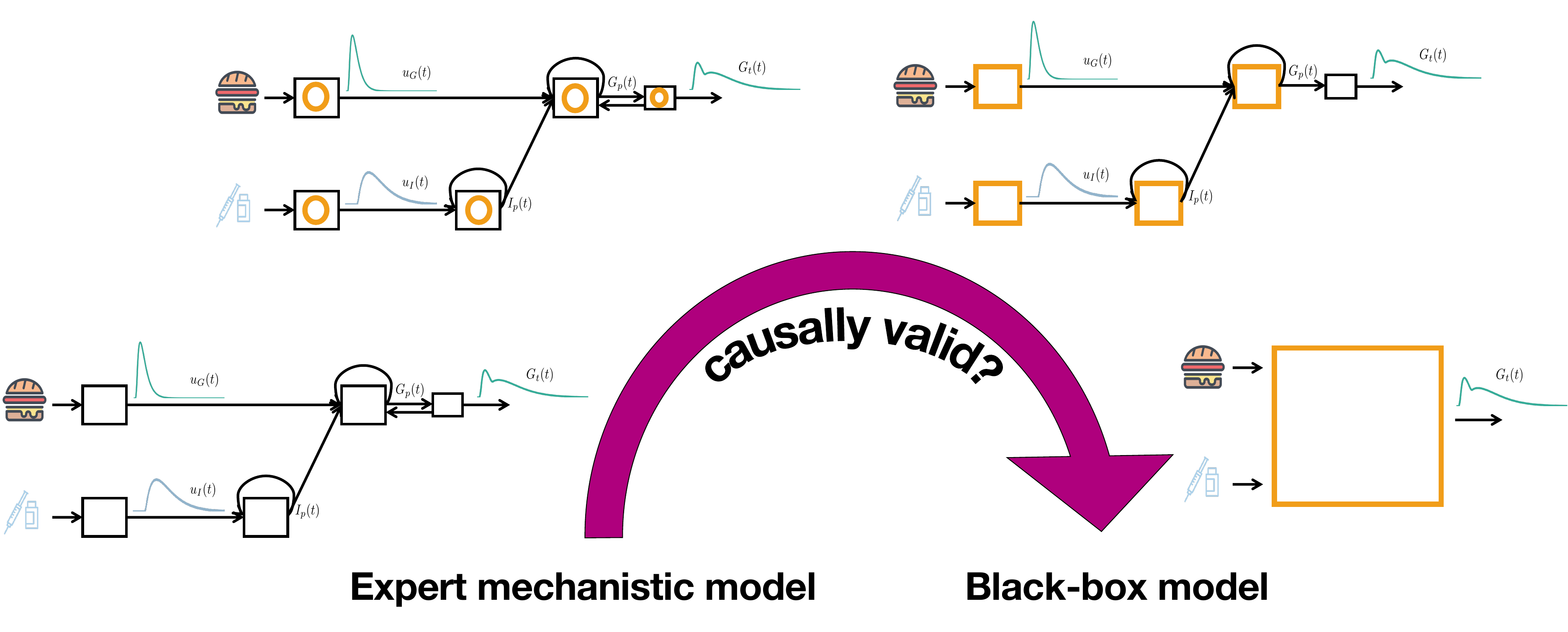}}
    \vspace{-0.1in}
    \caption{Visualization of hybrid modeling spectrum for a simple mechanistic model of insulin-glucose dynamics. From left to right: (1) mechanistic model; (2) latent parameter dynamics; (3) MNODE; (4) neural ODE. Dependencies between states are maintained until (4), with state dynamics increasingly flexible.}
    \label{fig:hybridspectrum}
    \end{center}
    \vskip -0.2in
\end{figure}

%
%Mechanistic models capture interpretable and causal dynamics, but are too restrictive for real-world data, while black-box models are highly flexible but prone to overfitting and loss of interpretability and causal validity. The search for reconciliation between mechanistic modeling and data-driven learning of complex dynamical systems has resulted in what is often referred to as \emph{hybrid} or \emph{graybox} modeling. There are two sides to the same coin of hybrid modeling: One is allowing more flexibility to our otherwise rigid mechanistic models; the other is adding (causal) inductive bias to our highly flexible neural ODE approach when domain knowledge is available~\cite{miller2021breiman}.  In either case, the result is to marry the ideas of mechanistic and neural modeling. 
In this section, we provide background on hybrid modeling. % approaches.
For %the sake of our 
this paper, it is useful to think of hybrid modeling as a spectrum from pure mechanistic whitebox models to fully blackbox approaches, as illustrated in Fig.~\ref{fig:hybridspectrum}.  There are two important components to the hybrid modeling spectrum: the degree to which neural networks learn the state dynamics and the degree to which the dependencies between states encoded in the mechanistic model are maintained. % associated with %\ebf{Ramesh, insert background on graphs here.}

The notion of a hybrid spectrum is an oversimplification as the space of possible hybrid models involves any number of combinations of hybridizations that are not easily ordered along a single axis.  However, this framework is helpful for pedagogical purposes. Further, the hybrid models we focus on in our experiments \emph{can} be ordered in terms of increasing model flexibility, which helps illustrate the risk that hybrid models can lose causal validity as we walk along this spectrum (see Fig.~\ref{fig:predictive-loss-results}). 

%\subsection{Hybrid Mechanistic Models}
%In this section, we review the foundational models we build upon in Section~\ref{sec:methods}.

\subsection{Mechanistic Models} On the far left-hand side of our hybrid modeling spectrum in Fig.~\ref{fig:hybridspectrum} is the pure mechanistic model specified via a set of ordinary differential equations (ODE),
\begin{align}
    \frac{ds}{dt} = m\bigl(s(t); \ \beta\bigr),
\end{align}
where $s$ is a vector representing the state, and $\beta$ is a vector representing the simulator parameters.
% Such mechanistic descriptions are common in a range of domains, including cardiac modeling \cite{davies2016recent},   drug metabolism \cite{johnson2010semi}, immunology \cite{brown2018applications},  epidemiology \cite{holmdahl2020wrong} and neural circuit \cite{ladenbauer2019inferring}. \ebf{include list and references}\bz{added}.
In applications where interventions or other external controls, $x$,
%---represented by a vector $x$---
are being applied to the observed dynamical system, we obtain a \emph{controlled} ordinary differential equation (CDE), 
%\rj{I deleted reference to dimension $d$ in the state vector, so that we don't need to explicitly mention the dimension for any of the other model components either.}
%
\begin{align}
    \label{eq:mechCDE}
    \frac{ds}{dt} = m\bigl(s(t),x(t); \ \beta\bigr).
\end{align}
For simplicity, we still refer to such systems as ODEs, just as controlled state space models are often simply called state space models.  Our focus will be on this controlled setting since our interest is in causal hybrid modeling to produce (valid) counterfactual simulations for an intervention $x$.\footnote{We use $x$ to represent both possible controls or interventions, as well as exogenous covariates relevant to the state dynamics.} In Fig.~\ref{fig:hybridspectrum}, we introduce a cartoon version of such a (controlled) ODE for insulin-glucose dynamics. 

%Sophisticated mechanistic models remain a preferred method in many domains, as they capture lab-validated or otherwise known physical or physiological causal properties of the system.  For example, the UVA/Padova simulator~\cite{man2014uva} we consider in this paper has been FDA approved as a substitute for pre-clinical trials in the development of artificial pancreas algorithms.  Of note, however, is that sophisticated mechanistic models tend to add states to capture more intricate, non-Markovian dynamics such as delays, while maintaining the tractability of linear ODEs, leading to over-parameterization, and as a result, nonidentifiability. %\delrj{over-parameterization, over-specification, \newmel{and practical unidentifiabilities}}. The latest version of the UVA/Padova simulator has over 30 states.  While highly parameterized, the solution space of these models fails to capture important dynamics observed in real-world data; moreover, their parameter inference is highly sensitive.
%\delrj{these models are brittle and tend to struggle to model noisy, real-world data collected under conditions that do not match the controlled laboratory or simulated environment settings.} 

\subsection{Neural ODE}
In the absence of mechanistic domain knowledge, neural ODEs~\citep{rico1992discrete,weinan2017proposal,haber2017stable,chen2018neural,kidger2022neural} have been proposed as a blackbox approach to learning such a dynamical system:
%\mel{For consistency w.r.t. $\beta$, I recommend $f(\cdot \ ; \ \theta)$ rather than $f_\theta$ throughout.}
\begin{equation}
    \frac{ds}{dt}=f(s(t),x(t);  \ \theta).
    \label{eq:NODE}
\end{equation}
Here, $f$ denotes a neural network with parameters $\theta$. Neural ODEs represent the far right side of the spectrum in Fig.~\ref{fig:hybridspectrum}. These methods have proven useful in many dynamical system modeling and simulation tasks, especially in the sciences where ODEs are a standard language for describing systems~\citep{qian2021integrating, lu2021neural, owoyele2022chemnode, asikis2022neural,li2022graph}. %\ebf{are there more references here to include?}

\subsection{Hybrid Models}
%\delebf{ \citet{levine2022framework} provide an outline of hybrid modeling in terms of four high-level approaches, the last three of which are also discussed in \cite{karniadakis2021physics}.
%  \rj{How are the three approaches we introduce below related to the three or four approaches in these two cited papers?  Some explanation is needed as a transition.}}
%\delrj{Instead of providing another comprehensive review, we restrict our attention here to the two basic building blocks we use in our Hybrid$^2$ modeling.}

%Hybrid models combine mechanistic models with the flexibility of neural networks. The hope is that hybrid methods not only allow one to solve complex modeling problems with improved precision and accuracy \cite{pathak2018hybrid,willard2022integrating}, but also reduce the demand on data \cite{rackauckas2020universal}, enhance reliability and robustness \cite{didona2015enhancing}, and make the machine learning algorithms interpretable \cite{karniadakis2021physics,du2019techniques}. Examples of successes of hybrid modeling in healthcare can be found in \cite{qian2021integrating,sottile2021real,hussain2021neural}.

For our purposes, we introduce a general hybrid model that can be used to represent several different sub-methods:
%\mel{I'm going to take a crack at having 1 equation to represent the different scenarios.}
\begin{equation}
\label{eq:hybridCDE}
\begin{aligned}
    \frac{ds}{dt} &= c_1 m\bigl(s(t),x(t); \ \beta(t) \bigr) + c_2 f_1\bigl(s(t),x(t), c_3 z(t);\ \theta \bigr) \\
    \frac{dz}{dt} &= f_2\bigl(x(t),z(t); \ \theta\bigr) \\
        \beta(t) &= \beta + c_4 f_3\bigl(x(t), z(t); \ \theta \bigr).
 %   \label{eqn:state-closure}
\end{aligned}
\end{equation}
This class of hybrid models can improve upon the nominal mechanistic model in Eq.~\eqref{eq:mechCDE} by introducing (i) time-varying parameters $\beta(t)$; (ii) a flexible correction term to the mechanistic ODE given by $f_1$; and (iii) a time-dependent latent state vector $z(t)$. We introduce constants $c_1, c_2, c_3, c_4 \in \{0,1\}$ to ``switch" between different sub-methods. If $c_1 = 1$ and $c_2 = c_3 = c_4 = 0$, we recover a fully mechanistic model. 
On the other hand, if $c_2 = 1$ and $c_1=c_3 = c_4 = 0$, we obtain a blackbox neural ODE.

For other choices of these constants, we obtain various previous hybrid methods.  When $c_1 = c_4 = 1$ and $c_2=c_3 =0$, we have a method that addresses infidelities of the mechanistic model solely via time-dependent parameters governed by latent dynamics $z$; we refer to this hybrid approach as {\em latent parameter dynamics}.  %\rj{This used to be called parameter closure; I like the new name better, as it matches up with my (evolving!) understanding of closure.}
In the context of glycemic modeling, \citet{miller2020learning} use a deep state space model to capture time- and context-dependence of the simulator parameters.
When $c_1 = c_2 = 1$, we learn a flexible correction $f_1$ to the mechanistic model in Eq.~\eqref{eq:mechCDE}.  We refer to this hybrid approach as {\em state closure}.  If we also have $c_3=1$, we learn a correction $f_1$ that is additionally dependent on latent states $z(t)$.  If instead we have $c_3=c_4=0$, we learn a simple closure model $f_1$ which only depends on the mechanistic states $s(t)$~\citep{rico1994continuous}.

%\rj{I rewrote the preceding paragraph a bit to make i clear that these are both state closure approaches; but I'm the least familiar with the definition of closure, so please check.  Also, we only ever have $c_1 = 1$ OR $c_3 = 1$.  Should we just set $c_3 = 1 - c_1$ and have one less parameter?  Is there ever a case where we could have a hybrid model with both $c_1 = c_3 = 1$?} \ebf{This is basically what we're doing with UVA closure...}

\citet{yin2021augmenting}, \citet{levine2022framework}, and \citet{karniadakis2021physics} provide related frameworks for defining the space of hybrid models. The above schemes, and those further described in~\citet{levine2022framework,karniadakis2021physics,yin2021augmenting}, are not mutually exclusive and are often combined~\citep{qian2021integrating,wu2022learning,wang2022causalgnn}.

Another important component of the mechanistic model that has been ignored to this point is the set of dependencies between states encoded by the mechanistic model.  To make these dependencies explicit, and to explore hybrid models that---at a dynamical modeling level---maintain the same dependencies, we introduce the notion of connectivity of the state dynamics via a set of adjacency matrices $A_s, A_x$:
%\delebf{We can also introduce additional causal structure into these equations by pre-defining the connectivity of the functions; we do this by introducing adjacency matrices $A_s, A_x$ and re-writing as}
\begin{equation}
\label{eq:hybridCausalCDE}
\begin{aligned}
    \frac{ds}{dt} &= c_1 m\bigl(s(t),x(t); \ A_s, A_x, \beta(t) \bigr) \\
& \ \ + c_2 f_1\bigl( s(t), x(t), c_3 z(t); \ A_s, A_x, \theta \bigr) \\
    \frac{dz}{dt} &= f_2\bigl(x(t),z(t); \ \theta\bigr) \\
        \beta(t) &= \beta + c_4 f_3\bigl(x(t), z(t); \ \theta \bigr).
\end{aligned}
\end{equation}
Here the $(i,j)$ entry of $A_s$ is 1 if $ds_i/dt$ is allowed to depend on $s_j(t)$, and similarly for $A_x$.  In this way, the adjacency matrices $A_s, A_x$ encode a structural equation model that constrains which state variables in $s, x$ are allowed to influence each component of $\frac{ds}{dt}$.  When $c_1 = c_2=1$, in contrast to the state closure model of Eq.~\eqref{eq:hybridCDE}, here the additive neural network correction is also forced to respect the dependencies between states.
%\rj{Matt got me on board with his adjacency matrix formulation :)  I added some additional explanation just to ensure it's clearly defined.  Note I also added $A_s$ and $A_x$ to $m$ to allow us to formulate the hybrid models in Section 3.}

%\mel{My GOAL here is to have fewer equations, and have methods be unified. My hope is that now the upcoming 5+ equations can essentially be removed and described with choices of $c_1, c_2, c_3, A_s, A_x$ in \eqref{eq:hybridCausalCDE}.}
% give which are determined by a function by latent dynamics
% has latent dynamics $z(t)$, time-varying mechanistic parameters $\beta$ that are governed by the latent dynamics $z$, and an explicit correction term on

%In Eq.~\eqref{eq:hybridCausalCDE}, we add an indicator variable $c_4$ which selects whether or not the form of the state dynamics defined in $m$ are used.
Note that in Eq.~\eqref{eq:hybridCausalCDE}, when $c_1 = 0$ but $c_2 = 1$, we maintain the dependencies between states (specified via $A_s$ and $A_x$), but allow the resulting state dynamics to be fully learned via neural
networks.  We refer to this model as a \emph{mechanistic neural ODE} (MNODE).  MNODE represents the third illustrated model in the hybrid spectrum of Fig.~\ref{fig:hybridspectrum}.

\section{Hybrid$^2$ Modeling}
\label{sec:methods}
Hybrid models offer significant promise in many scientific domains where ODE-based mechanistic models are commonly deployed: they can provide important inductive bias, interpretability, and causal grounding as compared to their black-box alternatives; and, compared to pure mechanistic modeling, hybrid models can reduce bias through the flexibility of neural network components.  However, as one walks along the hybrid modeling spectrum---from pure mechanistic to pure blackbox---these touted advantages can quickly disappear, as illustrated in Fig.~\ref{fig:predictive-loss-results}.  In particular, there is nothing constraining a hybrid model---trained on a predictive performance objective---to maintain the causal structure and interpretability of the original mechanistic model.  

We address this challenge in Sec.~\ref{sec:hybridloss} by introducing a \emph{hybrid loss} that mixes predictive performance with causal validity.  For the latter, we again lean on domain knowledge (as we did in our use of a mechanistic model), but this time cast as knowledge about the \emph{direction} of treatment effects.  In other words, we presume that we know in advance that applying, e.g., $2x$ instead of $x$ to the system will increase (or decrease) the value of a score function that depends on the observed dynamic (counterfactual) response.  %For example, in modeling glucose dynamics, we know that the more carbohydrates you consume, the more your glucose will rise and the more insulin you dose, the more your glucose will fall.  
Encoding this information in the loss function helps us achieve a win-win of increased modeling flexibility while biasing the learning towards solutions that match our causal knowledge.

\subsection{Limitations of Hybrid Models}
%One use of hybrid modeling, and the use case of interest in our clinical setting, is for robust counterfactual simulations to be used for downstream decision making.  However, while hybrid methods may be able to produce accurate predictions, the methods can struggle to identify causally coherent models, especially when considering the more flexible end of the hybrid modeling spectrum.  

To illustrate the potential for hybrid models to lose causal validity, take the MNODE model of Eq.~\eqref{eq:hybridCausalCDE} where $c_1 = 0$.  Relying solely on the mechanistic dependency graph between states while learning state dynamics can fail to distinguish parameters leading to correct directions of treatment effects. Consider data generated by a mechanistic ODE $ds/dt = -as+bx+c$, where $a,b,c$ are positive. Here, the rate of change of $s$ is monotonically increasing with %respect to 
$x$ and therefore an intervention that increases $x$ should have a positive treatment effect on $s$. However, after turning these dynamics into MNODE, we instead have $ds/dt = f(s,x; \theta);$
here, we lose the monotonicity with respect to $x$, unless a monotone neural network is intentionally used---which would severely constrain the flexibility of MNODE, the original motivation for adopting this modeling approach.

\begin{figure}[t!]
    %\vskip 0.2in
    \begin{center}
    \includegraphics[width=\columnwidth]{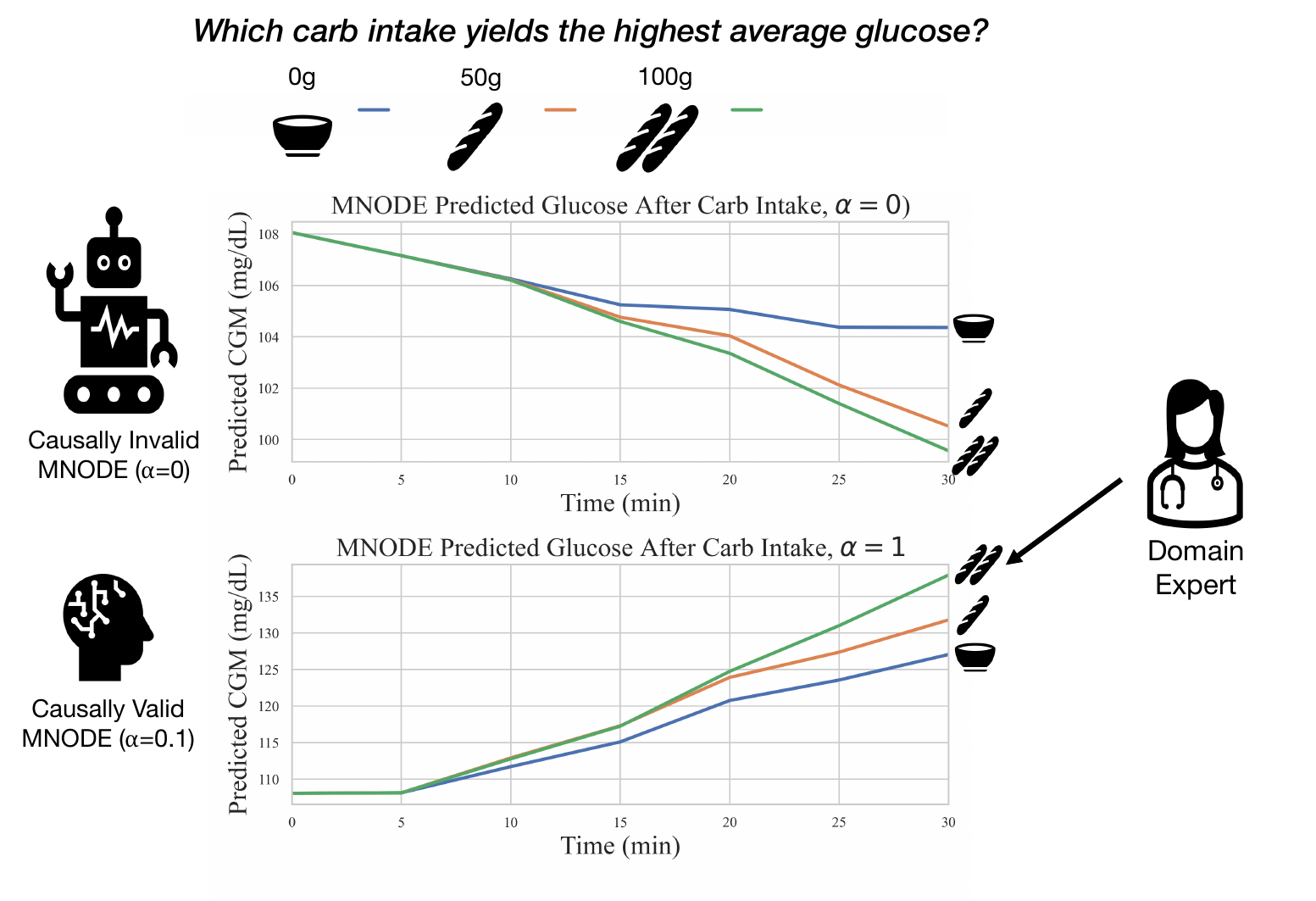}
    \vskip -0.15in
    \caption{Counterfactual simulations of 3 levels of carbohydrate intake: none (\textcolor{blue}{blue}), 50g (\textcolor{orange}{orange}), 100g (\textcolor{teal}{green}), comparing MNODE trained using predictive loss ($\alpha=0$) and a hybrid loss ($\alpha=0.1$). For $\alpha=0$, MNODE incorrectly selects ``none'' as the intervention causing the largest resulting average blood glucose, because it incorrectly infers that carb intake decreases glucose.  The model using $\alpha=0.1$ correctly selects ``100g'', consistent with causally valid domain knowledge ($\bestI=\estbestI$).}
    \label{fig:counterfactual-sim}
    \end{center}
    \vskip -0.2in
\end{figure}

One might argue that while not explicitly encoded in the architecture, neural networks should be able to learn relatively simple signals like monotonicity from the data. Unfortunately, this is not the case when learning from noisy, observational datasets.  For example, in T1D, the dosing of bolus insulin (negative effect on glucose) frequently occurs in close succession with a planned intake of carbohydrates (positive effect). Due to the confounding effect of insulin, the hybrid model may incorrectly infer that if an individual consumes carbohydrates, glucose drops; 
%\delrj{because the previously dosed insulin raises glucose} 
see Fig.~\ref{fig:counterfactual-sim}. We provide an exploration of this effect in Sec.~\ref{sec:results}. The resulting dilemma is, while we do not want to make simplifying assumptions like linearity and monotonicity, we want to encourage our hybrid models to learn dynamics and interactions consistent with known signs of treatment effects. %\footnote{\newrj{Another related note is that even the mechanistic models themselves, upon which our hybrid models are built, are restricted to the causal structure predefined by the ODE.  As such, these models may fail to correctly evaluate interventions in settings where either the causal mechanism breaks down or otherwise varies; for example, because the UVA/Padova simulator does not encode all activity-related mechanisms, it performs poorly in classifying interventions related to exercise mode and intensity--leading to nonzero classification error rate in Fig.~\ref{fig:predictive-loss-results}.}}
%For example, the impact of physical activity on blood glucose levels is not described in the UVA/Padova simulator, which limits the simulator's ability to infer the impact of various types of exercise. 
% As discussed in Sec.~\ref{sec:intro}, the UVA/Padova simulator does not encode all activity-related mechanisms. This is exactly why in Fig.~\ref{fig:predictive-loss-results}, even the UVA/Padova model and its hybrid variant with latent parameter dynamics do not achieve zero classification error: the set of potential interventions contains not only those involving carbohydrates and insulin dosage, but also those about exercise modes and intensities.}

%First, mechanistic and hybrid models are constructed to address causal relationships, and yet are also confined by the same underlying causal architecture, which restricts their ability to tackle causal queries not pre-defined within the mechanistic model. 

\subsection{The Hybrid Objective}
\label{sec:hybridloss}
%Just as we may have mechanistic domain knowledge in the form of an ODE, there are also many cases where we know the direction of treatment effects.    Such causal knowledge is not straightforward to encode in hybrid models themselves.  Instead,
Motivated by the preceding discussion, we encode causally relevant domain knowledge in hybrid models through the loss function itself, with the goal of biasing the learning towards causally valid models.

%In this section we first describe both predictive and causal losses, and then a hybrid loss that combines them.  We then describe how we use this hybrid loss for model training.

%\paragraph{Causal Task as Intervention Optimization} To encode our domain knowledge on the relative ranking for a set of 

%As discussed in the introduction, predicting the exact treatment effect can be unrealistic in many settings. We therefore relax the causal task we consider to what is most relevant in practice----picking the best intervention. 

\prepar
\paragraph{Preliminaries}
Let $X(\Tinterval) = \left\{ x(t); t \in \Tinterval\right\}$ be the set of control inputs at times in $\Tinterval$, $\Tinterval \subset \mathbb{R}$.  In our diabetes example, $x$ can represent interventions such as carbohydrate intake or insulin dosing for a single patient. In our preceding development, we described models for the evolution of the full state $[s(t);z(t)]$.  
Here we assume we have a partial observation of that state; for example, in our diabetes case study, we observe glucose measurements via a continuous glucose monitor (CGM).  We denote this partial state observation by $y(t)$ and assume without loss of generality that $y(t) = H[s(t);z(t)]$ for some projection matrix $H$ (which can be the identity matrix, $I$, if we have complete observations). We then define $Y(\Tinterval) = \left\{ y(t); t \in \Tinterval\right\}$ as the corresponding set of (partial) observations over $\Tinterval$.

We assume a context window $\Tcontext$ of controls and observations, $(X(\Tcontext),\allowbreak Y(\Tcontext))$.  We are interested in the behavior of $y(t)$ over a prediction window $\Tpred$, represented by $Y(\Tpred)$, given future controls $X(\Tpred)$.
A {\em model} $M$ takes as input $(X(\Tcontext),Y(\Tcontext)$, $X(\Tpred))$ and produces an output trajectory of $y$ over the prediction window.  Below we use $\trueM$ to represent the unknown ground truth model; this is the model that generated the observations $Y(\Tpred)$.  We use $\estM$ to denote a fitted model. Throughout we use $\estY(\Tpred) = \left\{\hat{y}(t); t \in \Tpred \right\}$ to denote our predicted output from such a fitted model.
%\newmel{We describe further below how we fit models using the losses we define.}

\prepar
\paragraph{Predictive Loss}
We evaluate the predictive loss of an estimated model using mean squared error:
\begin{align} L_{\pred}(\estM) = \sum_{t \in \Tpred} \Big\|y(t) - \hat{y}(t)\Big\|_2^2.
\label{eq:PredictiveLoss}    
\end{align}
In practice, we have a collection of observed sequences (e.g., one per patient) and compute the loss summing over all sequences.  We then normalize our loss by the total number of observed time points aggregated over the sequences.

\prepar
\paragraph{Causal Loss}  
To incorporate causal validity into the loss function, we develop a causal loss that examines \emph{counterfactual simulations} rather than \emph{predictions}. For fixed context $X(\Tcontext), Y(\Tcontext)$, we consider a range of hypothetical interventions $X^{(i)}(\Tpred)$, $i = 1, \ldots, K$, that differ from the observed $X(\Tpred)$.  %\rj{Is the word "synthetic" the best choice in the previous sentence, given that we use "synthetic" repeatedly to refer to our simulated data example?  Here we could say "hypothetical" or something similar} 
The ground truth model would produce observations $Y^{(i)}(\Tpred) = \trueM(X(\Tcontext), Y(\Tcontext), X^{(i)}(\Tpred))$ for the $i$'th intervention. % \newmel{that would differ from the observed $Y(\Tpred)$}.  
If we could observe these counterfactuals, then we could compute the {\em causal effect} of each intervention: % using the utility:
\[ \tau^{(i)} = u(Y^{(i)}(\Tpred)) - u(Y(\Tpred)).
\]
Here, $u$ is a {\em score function} that computes a meaningful scalar-valued output from a sequence input.
%\mel{Do we really want to maximize glucose levels in diabetes? This doesn't sound quite right...Maybe an oversimplification of avoiding lows during exercise?}
For example, in diabetes, this score might correspond to the average glucose level over $\Tpred$. In other settings, the relevant quantity may be related to extremes, in which case $u$ may compute, e.g., the maximum or minimum value of the sequence.  

In many settings with complex dynamics or only very partial observations, such as in modeling glycemic response, the latent sources of variation (e.g., sleep, stress, physiology, etc.) render attempts to precisely estimate treatment effects nearly impossible. Importantly, in our setting the clinically relevant quantity for guiding patient care is the ordering of interventions; the same would be true in a multitude of settings where the model is guiding decision making. This ordinal domain knowledge is often readily available; for example, we know that holding other context constant, increasing insulin dosing should lower glucose levels, and increasingly larger insulin doses lead to greater reductions in glucose levels.  Even such ordinal causal grounding is absent in purely predictive models.

%Domain knowledge provides valuable qualitative guidance on the {\em direction and ordering} of these treatment effects: for example, we know that holding other context constant, increasing insulin dosing should lower glucose levels, and more insulin leads to greater reductions.  Unfortunately, often these causal effects cannot be precisely quantitatively estimated.  For example, in managing diabetes, exact estimation of glucose levels is exceedingly challenging even in highly controlled settings.  

To bias our training towards models that capture this ordinal causal domain knowledge, we introduce a causal loss where we evaluate whether the model can indentify the intervention with the maximum score.
%Instead, in our causal loss, we evaluate whether the model can
%%make progress by %focusing instead on %encouraging the model to
%identify the intervention with the maximum value, rather than precise estimation of causal effects.  
We define:
\[ \bestI = \text{argmax}_i \tau^{(i)} = \text{argmax}_i u(Y^{(i)}(\Tpred)). \]
In principle, we can compare this intervention to an estimated maximum score intervention under the trained model $\estM$, i.e.,
\[ \estbestI = \text{argmax}_i u(\estY^{(i)}(\Tpred)), \]
where for the $i$'th intervention we have
%\revbz{where}
$\estY^{(i)}(\Tpred) = \estM(X(\Tcontext), Y(\Tcontext),$ $X^{(i)}(\Tpred))$.
%\revbz{for the $i$'th intervention.}  

Of course, to enable model training, we need a loss function that admits a gradient.  Accordingly, rather than computing the exact maximum score intervention under the estimated model, we compute the {\em softmax} vector:
\[ \estQ = \sigma\Big(\phi u\big(\estY^{(1)}(\Tpred)\big) \ , \ \ldots \ , \ \phi u\big(\estY^{(K)}(\Tpred)\big)\Big). \]
Here $\sigma$ is the softmax function and $\phi > 0$ is a scalar ``temperature" parameter: the larger $\phi$ is, the closer $\estQ$ is to approximating a one-hot encoding of the true argmax.  
% Matt: removed () from below sentence.
In practice we choose $\phi$ as large as possible while maintaining stable training.
We compute the cross entropy (CE) loss between $\estQ$ and a one-hot encoding of the true argmax $\bestI$, denoted $CE(\estQ, \bestI)$.  We refer to this as our {\em causal loss}:
\begin{equation}
    \label{eq:CausalLoss}
    L_\causal(\estM) = CE(\estQ, \bestI).
\end{equation}
Again, in practice, we average over all observed sequences, as well as all intervention scenarios under consideration.

If one instead wants to evaluate a loss that considers the entire ranking over interventions, then we can in principle evaluate the CE loss in Eq.~\eqref{eq:CausalLoss} for every possible subset of interventions.  Of course, this rapidly becomes intractable with increasing $K$; in future work we plan to directly consider a loss function on the estimated ranking from $\estM$.

\prepar
\paragraph{Hybrid loss} For $\alpha \in [0,1]$, we define our {\em hybrid loss} as:
\begin{equation}
\label{eq:HybridLoss}
    L_\hybrid(\estM) =  (1-\alpha)L_\pred(\estM) + \alpha L_\causal(\estM),
\end{equation}
where we have overloaded notation to interpret $L_\pred$ and $L_\causal$ as the average predictive and causal losses, respectively, over all sequences and scenarios.  Note that when $\alpha = 0$, model fitting focuses entirely on prediction; when $\alpha = 1$, model fitting focuses entirely on causal validity.

\prepar
\paragraph{Model Fitting: The \modelname Approach}
A critical question in fitting hybrid ODE models is the initial condition $(s_0=s(t_0),z_0=z(t_0))$, where $t_0$ is the beginning of our prediction window. Two primary approaches exist for performing initial state estimation jointly with hybrid ODE parameter learning: 
(1) 
% differentiable % filtering of  %[filtering and SSEM are ~synonmys here]
statistical state estimation methods~\citep{chen2022autodifferentiable,brajard2021combining,Ribera_2022,levine2022framework} and 
(2) blackbox encoder models~\citep{chen2018neural}. We take the latter approach, viewing initial state estimation through the lens of sequence-to-sequence (seq2seq) modeling. 
% \ebf{Matt, check this!  I grabbed the following from the results section and not sure how best to blend...just add Chen 2022 next to Levine 2022?} 
% \mel{Instead, \cite{levine2022framework} and \cite{chen2022autodifferentiable} leverage auto-differentiable sequential state-estimation techniques to implicitly initialize and filter dynamical states during gradient-based optimization of hybrid ODEs.}

In particular, we produce the predicted sequence $\estY(\Tpred)$ using $X(\Tpred)$ and the initial condition $(s_0,z_0)$ encoded from the context data $(X(\Tcontext),Y(\Tcontext))$.
%In this paper, 
We consider a general blackbox encoder $M_\context$ for the initial conditions:
\begin{equation}
    (s_0,z_0) = M_\context\big(X(\Tcontext),Y(\Tcontext);\theta_\context\big), \nonumber
\end{equation}
where $\theta_\context$ are the parameters of $M_\context$.

We take the decoder, $M_\pred$, to be a hybrid ODE that takes as input the initial condition $(s_0, z_0)$, controls $X(\Tpred)$, and parameters 
$\theta_\pred = \{\beta,\theta\}$, where $\beta$ are the simulator parameters and $\theta$ the neural network parameters.  We assume the hybrid model indicator variables $c_1,\dots,c_4$ and adjacency matrices $A_s,A_x$ are specified.  The predicted values are produced as the output of numerical integration of this hybrid ODE decoder over the window $\Tpred$:
\begin{equation}
    \hat{y}\big(t_k\big) = H \cdot \mbox{Integrate}(M_\pred, \Tpred; X(\Tpred), s_0,z_0,\theta_\pred). \nonumber
\end{equation}
Here, $H$ is an indicator matrix selecting the observed states. The parameters $\theta_\context$ and $\theta_\pred$ are jointly optimized to obtain a fitted model $\hat{M}$ that minimizes the hybrid loss of Eq.~\eqref{eq:HybridLoss}.  We refer to this approach of training such a model by optimization of hybrid loss as \modelname.\footnote{Our implementation of H$^2$NCM is available at \href{https://github.com/bobjz/H2NCM}{\color{blue}https://github.com/bobjz/H2NCM}.}

\section{Related Work}

\paragraph{Neural Causal Models (NCMs)}
NCMs use feed-forward neural networks to extend the flexibility of structural equation models (SEM) \citep{pearl1998graphs}.
% SEMs are formulated as $V=f(V_\textrm{pa}, U_\textrm{conf}; \ \theta)$, in which the state of random variable $V$
% is defined by a structural equation defined by $f(\cdot, \ \theta)$ for which $V_\textrm{pa}$ are causal parents and $U$ confounders.
% While traditional SEMs envisage $f$ as a pre-defined function with few tunable parameters $\theta$, NCMs instead treat $f$ as a neural network with many learnable parameters.
% is a random variable, $V_\textrm{pa}$ represen its (random) causal parents, and U a random variable representing unknown confounders 
% use feed-forward neural networks, like MLPs, to \mel{Maybe instead "represent the functional forms in an SEM" ? Not sure if that is a correct understanding.} replace structural equations in a structural equation model (SEM) \cite{pearl1998graphs}, as explored in \cite{xia2021causal}.
% \mel{okay--how, roughly? and in what context? $y = F_{NCM}(x)$ ?}
% dynamical systems with MLPs, focusing on deterministic systems represented by ODEs, in contrast to NCM's broader SEM framework. 
% }
% \mel{What is SEM? not defined.} 
Note that both NCM and MNODE face the flexibility-causality dilemma in observational data settings where causal inference is challenging.  Similar to the role of SEM in NCM, MNODE replaces system equations--here, dynamical systems defining mechanistic ODEs---with neural networks, while retaining the original state-connectivity graph encoding specific causal relationships.
\citet{xia2021causal} highlighted NCM's limitations in treatment effect estimation without strong assumptions, while we empirically show MNODE's inability to learn the ordering of treatment effects without causal constraints. For MNODE and related hybrid modeling approaches, we address this challenge through introducing a causal loss; in contrast, \citet{xia2022neural} rely on strong distributional assumptions to maintain causality. Similarities between our work may suggest that our hybrid loss can be applied to improve causal alignment of general NCMs. %\citet{zevcevic2021relating} attempts to blend GNNs with causal relations by implementing NCMs as GNNs.

\prepar
\paragraph{Physics-informed Neural Networks (PINNs)}
PINNs, introduced by \citet{raissi2019physics}, use neural networks for solving partial differential equations (PDEs) by approximating solutions that conform to known differential equations.  In contrast to the hybrid models of this work, PINNs: 1) typically focus on learning \emph{solutions} to PDEs (rather than a governing vector field), and 2) do so via regularization in which a physics-based loss encourages the learned solution to comply with known physical laws. This method is advantageous when the system's underlying physics are well-understood, but computationally intensive or costly to simulate directly. Related is the systems-biology-informed deep learning approach by \citet{yazdani2020systems}, which constrains the learned models 
to exhibit properties similar to pre-specified mechanistic ODEs.

%\paragraph{Kernel methods and Gaussian Process Models}
%\mel{TBC}

\prepar
\paragraph{Graph Network Simulator (GNS)}
GNS applies the message passing architecture of graph convolution networks (GCN) to physical system simulation~\citep{sanchez2018graph,sanchez2020learning,pfaff2020learning,wu2022learning,allen2023graph}. \citet{poli2019graph} connect GNS with blackbox neural ODEs (GNODE) \citep[see also][]{jin2022multivariate,bishnoi2022enhancing,
wu2022learning,allen2023graph,li2022graph}. MNODE focuses on causal relationships between features \emph{across} time, rather than between spatial interactions; as such, we use directed rather than undirected graphs.

% \citet{zevcevic2021relating} attempts to blend GNNs with causal relations by implementing NCMs as GNNs.

%Our approach aligns with efforts to integrate GNNs with causal models \cite{zevcevic2021relating}, positioning MNODE as a deterministic simulator version of these GNN-NCM hybrids.

%They and their recent variants \cite{jin2022multivariate,bishnoi2022enhancing,li2022graph} typically involve undirected graphs, while MNODE uses directed graphs for simulation. MNODE differs in its focus on causal relationships in feature space, rather than spatial interactions in GNS. This approach aligns with efforts to integrate GNNs with causal models \cite{zevcevic2021relating}, positioning MNODE as a deterministic simulator version of these GNN-NCM hybrids.

%GCNs and GNSs are also relevant, utilizing message passing for node updates \cite{kipf2016semi,pfaff2020learning,wu2022learning,allen2023graph}. GCNs typically involve undirected graphs, while MNODE and GNS, including recent advancements \cite{jin2022multivariate,bishnoi2022enhancing,li2022graph}, use directed graphs for simulation. MNODE differs in its focus on time series data and causal relationships in feature space, rather than spatial interactions in GNS. This approach aligns with efforts to integrate GNNs with causal models \cite{zevcevic2021relating}, positioning MNODE as a deterministic simulator version of these GNN-NCM hybrids.
\section{Experiments}
\label{sec:experiments}

\subsection{Motivation: Safely Managing Exercise in T1D}
%\subsection{The exercise dilemma for diabetes patients}

Type 1 diabetes (T1D) is an autoimmune condition characterized by the destruction of insulin producing beta-cells. To manage glucose concentrations and prevent diabetes-related complications, intensive insulin therapy through externally delivered sources (e.g., insulin pumps) is required. Regular physical activity and exercise lead to numerous health benefits such as increased insulin sensitivity, weight management, and improved psychosocial well-being. However, for individuals with T1D, due to the inability to rapidly decrease circulating insulin concentrations, exercise can also increase the risk of hypoglycemia. Despite technological advancements such as continuous glucose monitoring (CGM) and automated insulin delivery %(AID) 
systems, individuals with T1D still face significant challenges with managing glucose concentrations around exercise. Many adults with T1D are not meeting current exercise recommendations of at least 30-minutes of moderate-to-vigorous physical activity %(MVPA) 
per day~\citep{riddell2017exercise}, with fear of exercise-related hypoglycemia a leading barrier~\citep{brazeau2008barriers}. To address these challenges, there is a pressing demand for precise, reliable models that can predict individualized glycemic responses during and after exercise and encourage safe physical activity for all individuals with T1D. 

Past efforts on modeling glycemic response to physical activity have focused on developing more intricate mechanistic models~\citep{dalla2009physical,liu2018enhancing,deichmann2023new}, rather than devising a performant predictive model. Although glucose prediction using ML and, more recently, deep learning methods has received significant attention, a dearth of papers consider exercise periods~\citep{oviedo2017review} and the few that do do not perform well~\citep{hobbs2019improving,xie2020benchmarking,tyler2022quantifying}. We aim to leverage hybrid modeling to predict glucose concentrations of an individual with T1D in the first 30 minutes following physical activity, given historical context and expected future covariates. %We chose a 30-minute post-exercise prediction window because %\deldz{most recorded exercises in our dataset (see Sec.~\ref{sec:data}) last for around 30 minutes}
Exercise leads to increased insulin sensitivity post-exercise and this may in turn contribute to a heightened risk of hypoglycemia after activity. %Modeling exercise-period glucose levels is already a challenging problem and we want to distinguish this task from the (even harder) task of the post-exercise period.
The post-exercise period represents both a period of glycemic risk and one in which interventions can be readily applied; our goal is to enable better guidance during this period.

\subsection{Data Preparation}
\label{sec:data}

%\paragraph{Data Source} 
Our data come from the Type 1 Diabetes Exercise Initiative (T1DEXI) \citep{riddell2023examining}, which can be requested via \url{https://doi.org/10.25934/PR00008428}. This is a real-world study of exercise effects on 497 adults with T1D. Participants in the study were randomly assigned to %one of three types of study-specific exercise videos: 
aerobic, resistance, or interval exercise videos for a total of six sessions over four weeks. Participants were asked to self-report their food intake and exercise habits, while their insulin dosage and relevant physiological responses were recorded by corresponding wearable devices such as insulin pumps, CGM devices, and smart watches.   

%\paragraph{Cohort Selection} 
We select participants on open-loop pumps, which enables real-time recording of insulin dosing with levels not proactively adjusted (and thus correlated with proximal CGM readings). %This makes sure we get complete insulin data that are not directly correlated with proximal CGM readings. 
We also limit our cohort to participants under 40 years old and with BMI below 30 because these participants tend to exercise more regularly and thus represent the general physically-active T1D population better. 

% \paragraph{Pre-processing} 
For selected participants, we filter out exercises shorter than 30 minutes. Then, for each exercise instance, we extract the participant's metabolic history (basal/bolus insulin delivery, carbohydrate intake, heart rate, step count, and CGM readings) 4 hours before to 30 minutes after the end of exercise to form a 5-dimensional time series. The 4-hour context window was chosen because the effect of most bolus insulin (fast-acting insulin) lasts for around 4 hours. % (carbohydrate digestion only takes about 1-2 hours) 
It is common within the T1D community to consider this time range when historical context is relevant in decision making. Finally, we filter out exercise instances with missing heart rate values. We end up with 143 exercise instances from 78 participants. See the Appendix for further 
data preprocessing details.

\prepar
%\vspace{-0.05in}
\paragraph{Intervention Sets} 
Our causal loss relies on the introduction of intervention sets. Here, we consider sets of size $K=3$. For our training procedure, for each training post-exercise instance, we create 3 replicates of $(X(\Tcontext)$,$Y(\Tcontext))$ and append each replicate with an input sequence $X^{(i)}(\Tpred)$ from a set selected uniformly at random from the following categories: (1) adding 0/50/100 grams (g) of carbohydrates at the end of exercise; (2) adding 0/2.5/5.0 units of insulin uniformly throughout the first 30 minutes post-exercise; (3) no-change/50g carbs/10.0 units of insulin at the end of exercise; (4) replacing the real post-exercise heart rate trajectory with a prototypical trajectory seen in aerobic/resistance/interval training.
All of these modifications are relative to the observed inputs $X(\Tpred)$.
These intervention sets capture the most integral pieces of well-understood and well-researched domain knowledge, allowing us to compute a reliable true ranking. % and yet are often messed by standard black-box models, as illustrated in Figure 1. 
In particular, increasing levels of carbohydrates (resp., decreasing levels of insulin) progressively increase mean glucose; and increasing exercise intensity (as measured by heart rate) leads to progressively increasing mean glucose during the short exercise period being considered~\citep{aronson2019optimal,riddell2017exercise}.  For our evaluation, we consider counterfactual simulations defined similarly. For each test exercise instance, we create a set of three instances $(X(\Tcontext),Y(\Tcontext), X^{(i)}(\Tpred))$ used to generate a counterfactual simulation $\hat{Y}^{(i)}(\Tpred)$, $i=1,2,3$. The intervention sets defining $X^{(i)}(\Tpred)$ are again drawn uniformly at random. In all cases (train and test), we take score to be average glucose in the 30-minute prediction window.

%For each exercise instance, we form three copies to modify with one of the following sets of interventions, selected uniformly at random: (1) adding 0/50/100 grams (g) of carbohydrate at the beginning of exercise; (2) adding 0/2.5/5.0 units of insulin uniformly throughout the first 30 minutes of exercise; (3) no-change/50g carbs/10.0 units of insulin at the beginning of exercise; (4) replacing the real exercise heart rate trajectory with a prototypical trajectory seen in aerobic/resistance/interval training. These intervention sets capture the most integral pieces of well-understood and well-researched domain knowledge, allowing us to compute a reliable true ranking. % and yet are often messed by standard black-box models, as illustrated in Figure 1. 
%In particular, increasing levels of carbohydrates progressively increase mean glucose, increasing levels of insulin progressively decrease mean glucose, and increasing exercise intensity (as measured by heart rate) leads to progressively increasing mean glucose during the short exercise period being considered~\citep{riddell2017exercise}.

\subsection{Key Implementation Details}
\paragraph{Discretization}
There are currently two mainstream methods to obtain approximate solutions to hybrid ODE variants:  differentiate-then-discretize and discretize-then-differentiate~\citep{ayed2019learning}. We choose the latter for its simplicity and adaptability. This method approximates the hybrid ODE system with stacks of residual networks \citep{he2016deep} via a forward-Euler discretization scheme
%\mel{I added $\beta_{t+1}$ since it wasn't defined.}
\begin{equation}
    \begin{aligned}
        s_{t+1}=s_t+&\Delta t \big[c_1 m(s_t,x_t; \ A_s,A_x,\beta_t) \\
                              &+c_2 f_1( s_t, x_t, c_3 z_t; \ A_s,A_x,\theta)\big]\\
        z_{t+1}=z_t+&\Delta t f_2(s_t, x_t ; \ \theta) \\
        \beta_{t+1} = \beta + &c_4 f_3(x_{t+1}, z_{t+1}; \ \theta ),
    \end{aligned} \nonumber
\end{equation}
and then computes the gradient via backpropagation. Here, we use subscript instead of parentheses to indicate the transition from the continuous to discrete time. In our implementation, we set $\Delta t$ to be 5 minutes, which is the sampling rate of CGM readings in the T1DEXI data.

\prepar
\paragraph{Model Reduction} Instead of the highly parameterized UVA/Padova S2013 model~\citep{man2014uva}, our hybrid models build upon a reduced model with the aim of (1) reducing variance and improving generalization performance, and (2) reducing training time.  We applied a data-driven reduction method to the full UVA/Padova mechanistic model and its causal graph; see the Appendix. An empirical comparison to hybridizations of the full, non-reduced UVA/Padova model are in the Appendix. 
 The results illustrate that the flexibility of the neural networks in the hybrid models is a satisfactory replacement for many of the full mechanistic model compartments. %\rj{The previous sentence doesn't seem to fully capture the point we want to make regarding variance; how about the following? "Our results illlustrate that hybridization using the full model severely increases variance with no improvement in performance; this suggests the flexibility of the neural networks in the hybrid models is a beneficial replacement for many of the full mechanistic model compartments."}
%As this reduction method is not central to the main thesis of the paper, the details are in the Appendix.

\prepar
\paragraph{Model Variants and Evaluation Procedure}
We consider 3 hybrid models in the form of Eq.~\eqref{eq:hybridCausalCDE} with increasing amounts of flexibility: (1) latent parameter dynamics (LP) defined via $c_1=c_4=1,c_2=c_3=0$; (2) latent parameter dynamics plus state closure (LPSC) defined via $c_1=c_2=c_4=1,c_3=0$; and (3) mechanistic neural ODE (MNODE) defined via $c_2=1,c_1=c_3=c_4=0$.  In all of these cases, the adjacency matrices $A_s$ and $A_x$ are specified based on our reduced UVA/Padova mechanistic model. We implement MNODE by defining a neural network for each output dimension and masking appropriate inputs, though the methods of \citet{chen2024structured} could be adapted to our setting to provide a more efficient alternative.
%\newmel{While we implement MNODE by defining a neural network for each output dimension and masking appropriate inputs, \citep{chen2024structured} offers a more efficient implementation.}
% that identifies a masking scheme suited for learning a single network.
We also consider the blackbox neural ODE model (BNODE) of Eq.~\eqref{eq:NODE} and tune the latent state dimension.
% defined as in MNODE via $c_2=1,c_1=c_3=c_4=0$, but with full adjacency matrices $A_s=A_x=\mathbf{1}$ to allow arbitrary causal relationships to be learned. \ebf{Discussing correctness of this statement with Bob...}  \rj{I think we should just drop adjacency matrices for BNODE -- see equation 3; could we just say "...(BNODE) defined as in Equation 3"?} 
Finally, we provide a set of non-ODE-based blackbox sequence models: the LSTM, Transformer (Trans), temporal convolutional network (TCN), and diagonal approximation to S4 (S4D)~\citep{gu2022parameterization}. For our mechanistic model baseline, we consider the full (not reduced) UVA/Padova S2013 simulator (UVA). %\newebf{A comparison to a reduced UVA/Padova model---which performs significantly worse---is provided in Appendix~\ref{Appendix:full-models}.} 
We limit all models to have fewer than 25,000 parameters for fair comparison. For all of our ODE-based models, we use a multi-stack LSTM for our initial-condition encoder; the non-ODE baseline models also take the historical context as input, with model-specific encoding choices.  See the Appendix for further model implementation details.

To tune our models and compute test error, due to the small dataset size, we use repeated nested cross validation (CV) with 3 repeats, 6 outer folds and 4 inner folds. The inner folds are used to tune hyperparameters and outer folds to estimate generalization error; results are averaged over three runs for which the sequences are randomly shuffled. For more details of the evaluation procedure, see the Appendix.

\subsection{Results}
\label{sec:results}

\begin{figure}[t!]
    %\vskip 0.2in
    \begin{center}
    \includegraphics[width=0.95\columnwidth]
{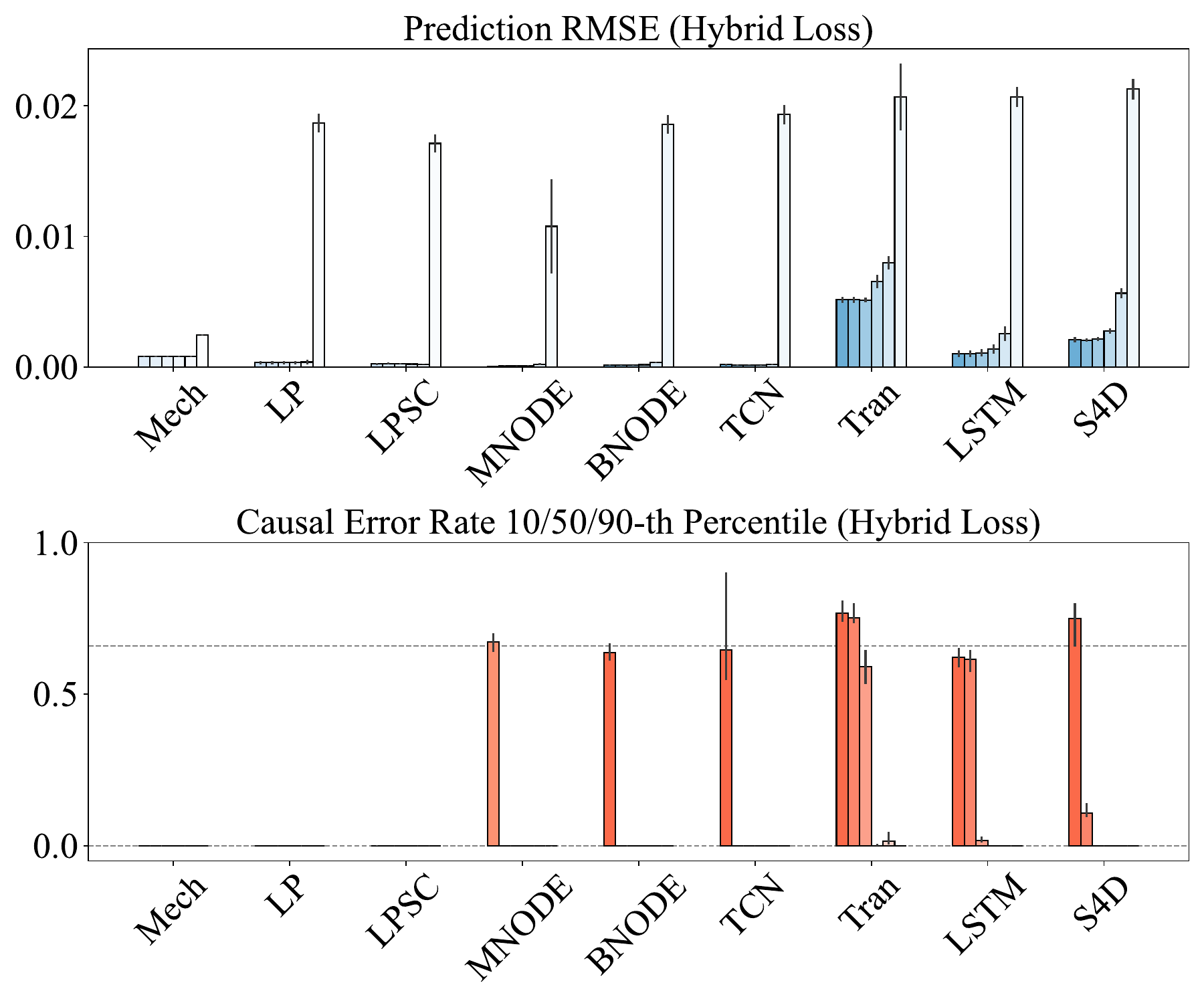}
    %{sections/img/alphas_pred_and_causal_synthetic.png}
    \vspace{-0.2in}
    \caption{For the synthetic data, predictive loss (RMSE) and standard error (\emph{top}) and 10/50/90-th(upper/bar/lower) percentile causal classification error rate (\emph{bottom}).  Within each bar group for a single model, % (Mech, LP, LPSC, MNODE, BNODE, TCN, LSTM, Tran, S4D), 
    the parameter $\alpha$ (hybrid loss parameter) increases from left to right, taking the values 
    $\{0, 1\mathrm{e}{-4}, 1\mathrm{e}{-3}, 1\mathrm{e}{-2}, 1\mathrm{e}{-1}, 1\}$.  The dashed line in the bottom figure corresponds to the causal classification error of random guessing ($2/3$).}
        \label{fig:synthetic}
    \end{center}
    \vspace{-0.25in}
\end{figure}

\paragraph{Synthetic Data}
We start by exploring a synthetic data example generated by a simple, single-state ODE with two correlated inputs, mimicking the carb/insulin correlation in the T1DEXI data.  In particular, the ODE is specified as:
\begin{align}
 dy/dt &= -y(t) + x_1(t) - x_2(t), \quad y(0) = 0\nonumber
 \end{align}
 \begin{align}
x_1(t) &= a\exp(-b t), \; %\nonumber \\
%& \quad 
a \sim \mbox{Unif}[1,2], \; b \sim \mbox{Unif}[5,15]\nonumber\\
x_2(t) &= 1.5 x_1(t)+\epsilon, \quad \epsilon \sim N(0,1\mathrm{e}{-4})\nonumber.
\end{align}
%where $\epsilon$ is zero mean noise. 
We simulate 600 training, 200 validation, and 200 test sequences and corresponding intervention sets, each selected uniformly at random from: (1) raising $x_1$ by 0/+1/+2; (2) raising $x_2$ by 0/+1/+2; or (3) no-change/adding +1 to $x_1$/adding +1 to $x_2$. The sequences are of length 100.

In Fig.~\ref{fig:synthetic}, we show prediction root mean squared error (RMSE) and classification error, calculated using the nested CV procedure described above, for each of the considered models as a function of $\alpha$. Here, the classification task is to correctly predict
the intervention that yields the highest score (taken to be average of $\hat{Y}^{(i)}(\Tpred)$) among the set of 3 choices. We see that even when the true mechanistic model is used in the hybrid modeling---and even when that system is very simple---our causal loss is critical in disambiguating the sign of the treatment effects for $x_1$ and $x_2$. For example, when $\alpha = 0$ (fully predictive loss), MNODE (and the blackbox models) have large causal loss because of lack of identifiability between the sign of the treatment effects of $x_1$ and $x_2$. However, even a small $\alpha>0$ rapidly enables disambiguation between $x_1$ and $x_2$. These synthetic results illustrate the importance of moving beyond predictive loss, especially when there is significant correlation between inputs; the importance is heightened when working with partial observations, as in our real-data setting below.

\begin{figure}[t!]
    %\vskip 0.2in
    \begin{center}
    \includegraphics[width=0.95\columnwidth]
    {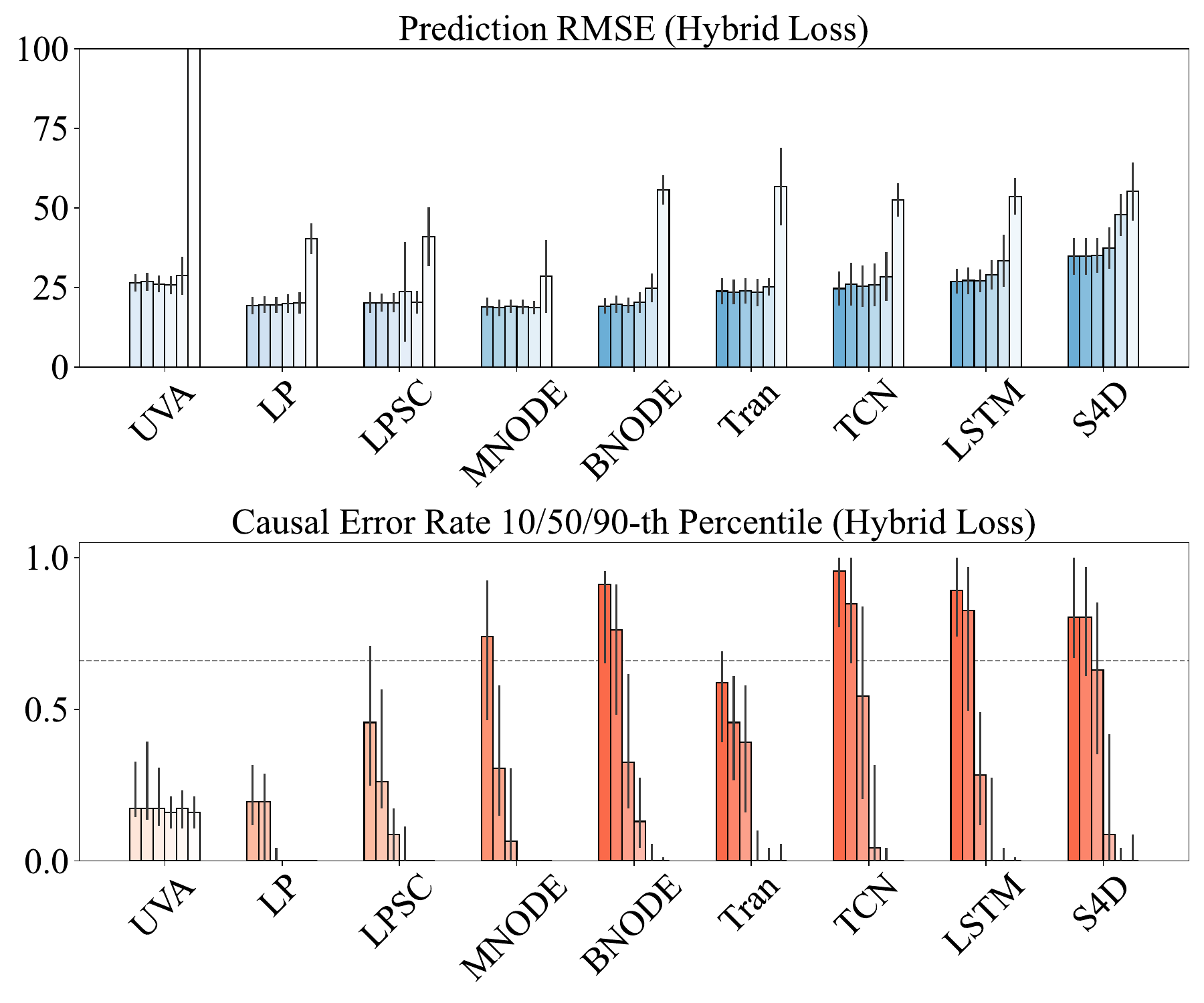}
    %{sections/img/alphas_pred_and_causal.png}
    \vspace{-0.2in}
    \caption{As in Fig.~\ref{fig:synthetic}, but for T1DEXI data and using a full UVA/Padova mechanistic model baseline and hybridizations of the reduced UVA/Padova simulator for LP, LPSC, and MNODE.}
    %\caption{\revbz{Predictive loss (RMSE) with its standard error and 10/50/90-th(upper/bar/lower) percentile} causal classification error rate for all models, across different values of $\alpha$ (hybrid loss parameter).  Within each bar group for a single model (UVA, LP, LPSC, MNODE, BNODE, TCN, \revbz{LSTM, Tran,} S4D), the parameter $\alpha$ increases from left to right as color changes from dark to light, taking the values $\{0, 1\mathrm{e}{-4}, 1\mathrm{e}{-3}, 1\mathrm{e}{-2}, 1\mathrm{e}{-1}, 1\}$ successively.  The dashed line in the bottom figure corresponds to the causal classification error of random guessing \revbz{($2/3$)}.}
        \label{fig:alphas}
    \end{center}
    \vspace{-0.25in}
\end{figure}

\paragraph{T1DEXI Data}
We now turn to our results on the real-world T1DEXI data. In Fig.~\ref{fig:alphas}, we see that as $\alpha$---the amount we emphasize the causal loss component of our hybrid loss---increases, all hybrid and blackbox models have significant \emph{decreases} in classification error; however, the hybrid causal error rate decreases more rapidly.  Critically, across a broad range of $\alpha$ values, the hybrid models have stable RMSE; the blackbox models appear to be more sensitive to increasing $\alpha$. (The RMSE of all models is impacted for sufficiently large $\alpha$, notably $\alpha=1$ when a purely causal loss is considered.) Note that the UVA/Padova mechanistic model does not achieve zero classification error because the mechanistic model does not include all mechanistic components relevant to exercise.  %For our hybrid ODEs, we see that for $\alpha \in \{0.01, 0.1\}$, we achieve very low classification error while maintaining low prediction RMSE.  
Our hybrid loss provides a win-win for hybrid ODEs: predictive performance exceeding pure mechanistic or blackbox approaches while also having extremely low classification error, again outperforming both the mechanistic and blackbox baselines.

Although our causal loss focuses on intervention \emph{ranking}, this information alone may bias learning towards causally-grounded models that better capture the direction of treatment effects even outside the domain knowledge encoded in the causal loss.  For example, in the Appendix, we define our causal loss in terms of rankings on carbohydrate and insulin intervention sets, but evaluate on insulin-to-carbohydrate ratios, which represents the most common intervention considered by clinicians for T1D. Our experiments demonstrate that even when we train with limited domain knowledge (i.e., the impact of carbohydrates or insulin individually), all models appear to exhibit improved insulin-to-carbohydrate rankings relative to models that do not leverage causal loss.

In the Appendix, we also consider the impact of increasing levels of corrupted domain knowledge where our causal loss is given the incorrect label for the top-ranked intervention, $\bestI$, for some fraction of intervention sets.  The results indicate robustness to moderate amounts of corruption.

%Recall, predicting glucose trajectories during exercise is a very challenging task.  Arriving at a highly performant predictive model that also enables valid counterfactual reasoning is thus an important step towards better guiding individuals with T1D to exercise safely.

\section{Discussion}
We presented \modelname, a method for integrating domain knowledge not only through a hybrid \emph{model}, but also through a hybrid \emph{loss} that encourages causal validity.  We consider this in the challenging real-world setting of predicting post-exercise glycemic response in T1D.  Our experiments illustrate a win-win where---across a wide range of settings of $\alpha$---our state-of-the-art predictive performance does not drop while the causal validity dramatically improves. In theory, models that can do both tasks well should align better with the true underlying system $\trueM$ and thus generalize better to unseen data, especially in the presence of distribution shifts, noise, and incompleteness, as our promising initial results in the Appendices suggest.
%Our experiments in the challenging real-world setting of predicting glucose responses during exercise in individuals with T1D demonstrate the importance of our \modelname approach.  We see
 %Future work will explore using a ranking error in place of cross entropy.  %(, all frequent situations in health datasets

We assume a known mechanistic model from which to build our hybrid models. One could instead consider softer forms of prior mechanistic knowledge. For example, \citet{wang2023neural} specify priors over graphical structures in neural differential equations. Indeed, our hybrid loss may be formulated in a Bayesian context as well, and could aid other modeling frameworks~\citep[cf.,][]{takeishi2021physics}.

%\newmel{The challenging features of real-world health data make unique identification of} \newebf{mechanistic} \newmel{models nearly impossible. 
%Uncertainty quantification can help cope with these difficulties by identifying distributions of likely models; Bayesian approaches, in particular, offer opportunities to incorporate softer forms of prior mechanistic knowledge. For example, \citet{wang2023neural} specifies priors over graphical structures in neural differential equations, which can be viewed as a valuable relaxation of our MNODE approach.Indeed, our hybrid loss may be formulated in a Bayesian context as well, and could aid other modeling frameworks , e.g., \citep{takeishi2021physics}.
%, both areas of future exploration.}

\section*{Acknowledgements}

\revbz{This work was supported in part by AFOSR Grant FA9550-21-1-0397, ONR Grant N00014-22-1-2110, NSF Grant 2205084, the Stanford Institute for Human-Centered Artificial Intelligence (HAI), the Helmsley Charitable Trust, and the Eric and Wendy Schmidt Center at the Broad Institute of MIT and Harvard. EBF is a Chan Zuckerberg Biohub – San Francisco Investigator.
This publication is based on research using data from Jaeb Center for Health Research Foundation that has been made available through Vivli, Inc. Vivli has not contributed to or approved, and is not in any way responsible for, the contents of this publication.}  We thank Ke Alexander Wang for helpful discussions on the implementation and training of various models.

\section*{Impact Statement}
% Authors are required to include a statement of the potential broader impact of their work, including its ethical aspects and future societal consequences. This statement should be in a separate section at the end of the paper (co-located with Acknowledgements, before References), and does not count toward the paper page limit. In many cases, where the ethical impacts and expected societal implications are those that are well established when advancing the field of Machine Learning, substantial discussion is not required, and a simple statement such as: 
%This paper presents work whose goal is to advance the field of Machine Learning. There are many potential societal consequences of our work, none which we feel must be specifically highlighted here.
All clinical decision support systems come with some risks, even with a trained human in the loop. If our intervention-ranking system gains the trust of the clinician, they may rely on its recommendations blindly. There are of course cases where it will make errors. In the type 1 diabetes setting, an error of recommending too much insulin may lead to potentially life-threatening hypoglycemia. A benefit of our system, however, is that instead of searching over all possible interventions, we are ranking an expert-provided list of possible interventions, all of which are assumed to be safe alternatives that would not deviate from clinically established standards of care.

% bibliography
\bibliography{refs}
\bibliographystyle{icml2024}

% APPENDIX
\newpage
\appendix
\onecolumn

\section{UVA-Padova Simulator S2013}
\label{uva-appendix}
Here we provide the exact full UVA-Padova S2013 model equations. Variables that are not given meaningful interpretations are model parameters.
\subsection{Summary Diagram}
\begin{figure}
    \centering
    \includegraphics[width=\linewidth/2]{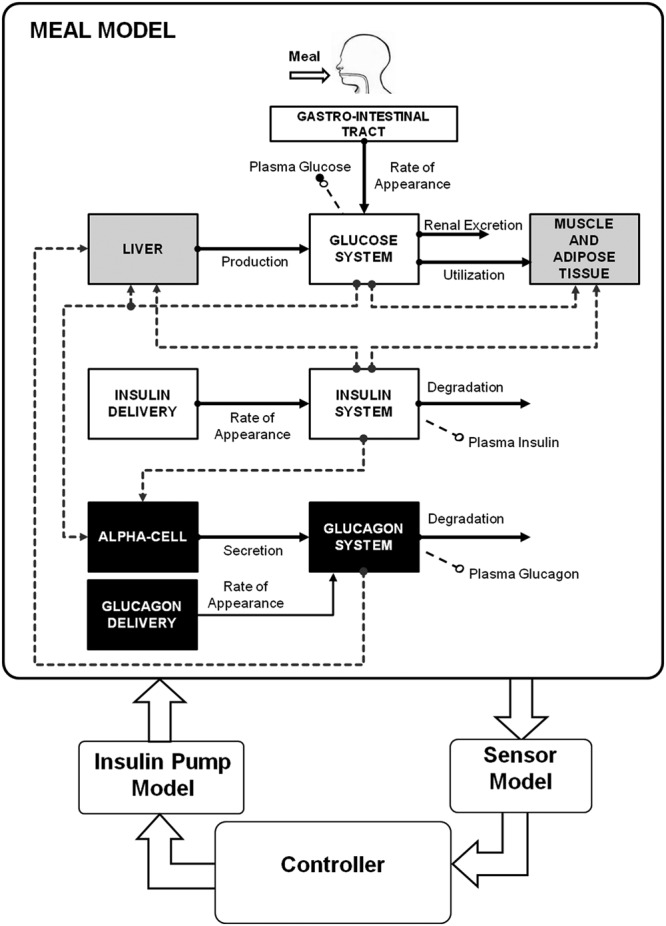}
    \caption{UVA/Padova Simulator S2013, taken from Figure 1 of \citet{man2014uva}}
    \label{fig:uvadiagram}
\end{figure}

At a high level, UVA-Padova can be summarized by the diagram in Figure \ref{fig:uvadiagram}, which is taken from Figure 1 in \citet{man2014uva}. It divides the complex physiological system into 10 subsystems, which are linked by key causal states such as Rate of Appearance, Endogenous Glucose Production and Utilization. Next, we will introduce each subsystem one by one and also explain the physiological meanings behind state variables.

\subsection{Glucose Subsystem}
\begin{align}
    &\dot{G}_p=EGP+Ra-U_{ii}-E-k_1G_p+k_2G_t\\
    &\dot{G}_t=-U_{id}+k_1G_p-k_2G_t\\
    &G=G_p/V_G
\end{align}
$G_p$: Plasma Glucose, $G_t$ Tissue Glucose, $EGP$: Endogenous Glucose Production Rate, $Ra$ Rate of Glucose Appearance, $U_{ii}$: Insulin-independent Utilization Rate, $U_{id}$: Insulin-dependent Utilization Rate, $E$ Excretion Rate, $V_G$ Volume Parameter, $G$ Plasma Glucose Concentration
\subsection{Insulin Subsystem}
\begin{align}
    &\dot{I}_p=-(m_2+m_4)I_p+m_1I_l+Rai\\
    &\dot{I}_t=-(m_1+m_3)I_t+m_2I_p\\
    &I=I_p/V_I
\end{align}
$I_p$ Plasma Insulin, $I_l$ Liver Insulin, $Rai$ Rate of Insulin Appearance, $V_l$ Volume Parameter, $I$ Plasma Insulin Concentration
\subsection{Glucose Rate of Appearance}
\begin{align}
    &Q_{sto}=Q_{sto1}+Q_{sto2}\\
    &\dot{Q}_{sto1}=-k_{gri} Q_{sto1}+D\cdot \delta\\
    &\dot{Q}_{sto2}=-k_{empt}(Q_{sto})\cdot Q_{sto2}+k_{gri}Q_{sto1}\\
    &\dot{Q}_{gut}=-k_{abs}Q_{gut}+k_{empt}(Q_{sto})\cdot Q_{sto2}\\
    &Ra=fk_{abs}Q_{gut}/(BW)\\
    &k_{empt}(Q_{sto})=k_{min}+(k_{max}-k_{min})(\tanh{}(\alpha Q_{sto}-\alpha b D)-\tanh{}(\beta Q_{sto}-\beta c D)+2)/2
\end{align}
$Q_{sto1}$: First Stomach Compartment, $Q_{sto2}$: Second Stomach Compartment, $Q_{gut}$: Gut Compartment, $\delta$ Carbohydrate Ingestion Rate
\subsection{Endogenous Glucose Production}
\begin{align}
    &EGP=k_{p1}-k_{p2}G_p-k_{p3}X^L+\xi X^H\\
    &\dot{X}^L=-k_i(X^L-I_r]\\
    &\dot{I}_r=-k_i(I_r-I)\\
    &\dot{X}^H=-k_HX^H+k_H\max(H-H_b)
\end{align}
$X^L$: Remote Insulin Action on EGP, $X^H$: Glucagon Action on EGP, $I_r$ Remote Insulin Concentration, $H$ Plasma Glucagon Concentration, $H_b$: Basal Glucagon Concentration Parameter
\subsection{Glucose Utilization}
\begin{align}
    &U_{ii}=F_{cns}\\
    &U_{id}=\frac{(V_{m0}+V_{mx}X(1+r_1\cdot risk))G_t}{K_{m0}+G_t}\\
    &\dot{X}=-p_{2U}X+p_{2U}(I-I_b)\\
    &risk=\begin{cases} 0 &G_b \leq G\\ 
                        10(\log(G)-\log(G_b))^{2r_2} &G_{th}\leq G<G_b \\
                        10(\log(G_{th})-\log(G_b))^{2r_2} &G<G_{th}\\\end{cases}
\end{align}
$F_{cns}$: Glucose Independent Utilization Constant, $X$: Insulin Action on Glucose Utilization, $I_b$ Basal Insulin Concentration Constant, $risk$ Hypoglycemia Risk Factor, $G_b$ Basal Glucose Concentration Parameter, $G_{th}$ Hypoglycemia Glucose Concentration Threshold.
\subsection{Renal Excretion}
\begin{equation}
    \dot{E}=k_{e1}\max(G_p-k_{e2},0)
\end{equation}
\subsection{Subcutaneous Insulin Kinetics}
\begin{align}
    &Rai=k_{a1}I_{sc1}+k_{a2}I_{sc2}\\
    &\dot{I}_{sc1}=-(k_d+k_{a1})I_{sc1}+IIR\\
    &\dot{I}_{sc2}=k_dI_{sc1}-k_{a2}I_{sc2}
\end{align}
$I_{sc1}$: First Subcutaneous Insulin Compartment, $I_{sc2}$: Second Subcutaneous Insulin Compartment, $IIR$ Exogenous Insulin Delivery Rate
\subsection{Subcutaneous Glucose Kinetics}
\begin{equation}
    \dot{G}_s=-T_sG_s+T_sG
\end{equation}
$G_s$: Subcutaneous Glucose Concentration
\subsection{Glucagon Secretion and Kinetics}
\begin{align}
    &\dot{H}=-nH+SR_H+Ra_H\\
    &SR_H=SR^s_H+SR^d_H\\
    &\dot{SR}^s_H=\begin{cases} -\rho\left[SR_H^s-\max \left(\sigma_2(G_{th}-G)+SR_H^b,0\right)\right]&\quad G\geq G_b\\
                                -\rho\left[SR_H^s-\max \left(\frac{\sigma(G_{th}-G)}{I+1}+SR_H^b,0\right)\right]&\quad G< G_b \end{cases}\\
    &\dot{SR}^d_H=\eta  \max(-\dot{G},0)
\end{align}
$SR_H^s$: First Glucagon Secretion Compartment, $SR_H^d$: Second Glucagon Secretion Compartment, $SR_H^b$: Basal Glucagon Secretion Parameter, $Ra_H$: Rate of Glucagon Appearance
\subsection{Subcutaneous Glucagon Kinetics}
\begin{align}
    &\dot{H}_{sc1}=-(k_{h1}+k_{h2})H_{sc1}+H_{inf}\\
    &\dot{H}_{sc2}=k_{h1}H_{sc1}-k_{h3}H_{sc2}\\
    &Ra_H=k_{h3}H_{sc2}
\end{align}
$H_{sc1}$: First Subcutaneous Glucagon Compartment, $H_{sc2}$: Second Subcutaneous Glucagon Compartment, $H_{inf}$ Subcutaneous Glucagon Infusion Rate.

\section{Reduced UVA-Padova for DTDSim2}
\label{app:ruva}
We used the a reduced version of UVA-Padova S2013 in the implementation of Latent Parameter Model and Latent Parameter with State Closure Model. It comprises the following equations:
\begin{align}
    &\dot{G}_p=EGP+Ra-U_{ii}-k_1G_p+k_2G_t\\
    &\dot{G}_t=-U_{id}+k_1G_p-k_2G_t\\
    &\dot{I}_p=-(m_2+m_4)I_p+m_1I_l+IIR\\
    &\dot{I}_l=-(m_1+m_3)I_l+m_2I_p\\
    &Q_{sto}=Q_{sto1}+Q_{sto2}\\
    &\dot{Q}_{sto1}=-k_{gri} Q_{sto1}+D\cdot \delta\\
    &\dot{Q}_{sto2}=-k_{empt}(Q_{sto})\cdot Q_{sto2}+k_{gri}Q_{sto1}\\
    &\dot{Q}_{gut}=-k_{abs}Q_{gut}+k_{empt}(Q_{sto})\cdot Q_{sto2}\\
    &Ra=fk_{abs}Q_{gut}/(BW)\\
    &k_{empt}(Q_{sto})=k_{min}+(k_{max}-k_{min})(\tanh{}(\alpha Q_{sto}-\alpha b D)-\tanh{}(\beta Q_{sto}-\beta c D)+2)/2\\
    &EGP=k_{p1}-k_{p2}G_p-k_{p3}X^L\\
    &\dot{X}^L=-k_i(X^L-I_p)\\
    &U_{ii}=F_{cns}\\
    &U_{id}=\frac{(V_{m0}+V_{mx}X)G_t}{K_{m0}+G_t}\\
    &\dot{X}=-p_{2U}X+p_{2U}I_p\\
\end{align}
Comparing to the full model, we performed the following:
\begin{enumerate}
    \item We replaced states $G,I$ with $G_p,I_p$ as they only differ by a constant.
    \item We removed the whole glucagon system to reduce model variance because most T1D patients do not have exogenous glucagon delivery, and their body's own glucagon regulation system is often impaired, a common symptom of T1D \citep{bisgaard2021mini}. 
    \item We removed the renal excretion system, as renal excretion of glucose only takes place during episodes of severe hyperglycemia, which does not happen very often to patients on insulin pumps. 
    \item We removed the subcutaneous glucose/insulin kinetics systems/the remote insulin state they are solely meant to introduce delays, which already exist in the data (CGM readings are only taken every 5 minutes).
    \item We removed the hypoglycemia risk factor $risk$ as it is used to model a relatively uncommon phenomenon.
\end{enumerate}
We verified these reduction changes on a validation set that was taken out of the training set to make sure they do not break our models.

\section{MNODE Graph Reduction Heuristic}
\label{app:graphred}
\begin{figure}[h]
    \centering
    \includegraphics[width=\linewidth/2]{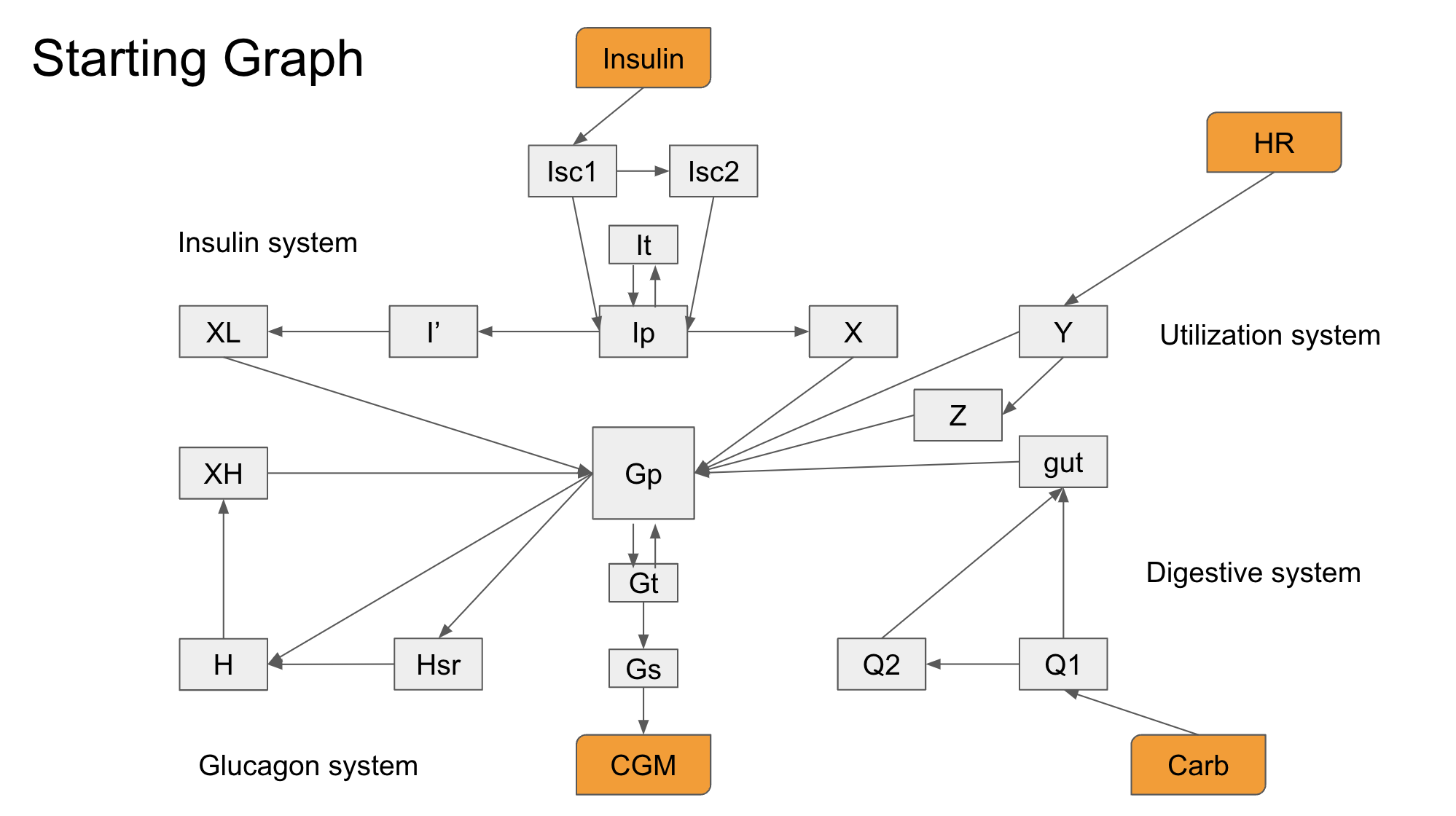}
    \caption{The Starting Graph for MNODE Graph Reduction Heuristic}
    \label{fig:startinggraph}
\end{figure}

We start with the full UVA/Padova causal graph as shown in Figure \ref{fig:startinggraph}, this graph is obtained by adding the physical activity model graph from \citet{dalla2009physical} to the UVA/Padova S2013 \citep{man2014uva} causal graph. Note that in the illustration, the node HR actually refers to both heart rate and step count, as we consider these two features both crucial indicators of physical activity intensity.
\begin{enumerate}
    \item Step 1: For each Strongly Connected Component in the graph (indicated in Figure {\ref{fig:showscc}}), try collapsing it into a single node and evaluate MNODE's performance with the resulting graph on the validation set. Adopt the change that (1) has the least loss among all trials and (2) has loss that is within 10 percent increase of the best loss ever achieved so far. The second condition is to make sure we are not picking among a set of bad choices and at the same time to encourage exploration (i.e. can proceed as long as loss does not increase too much). The metric we use is the hybrid loss with $\alpha=0.6$ (we picked an $\alpha$ that is not used in the actual experiments to avoid bias). 
    \item Step 2:  Repeat step 1 until no change satisfy both criteria. Figure \ref{fig:sccstep} shows an illustrative summary of step 1 and step 2.
   \begin{figure}
     \centering
     \begin{subfigure}[b]{0.3\textwidth}
         \centering
         \includegraphics[width=\textwidth]{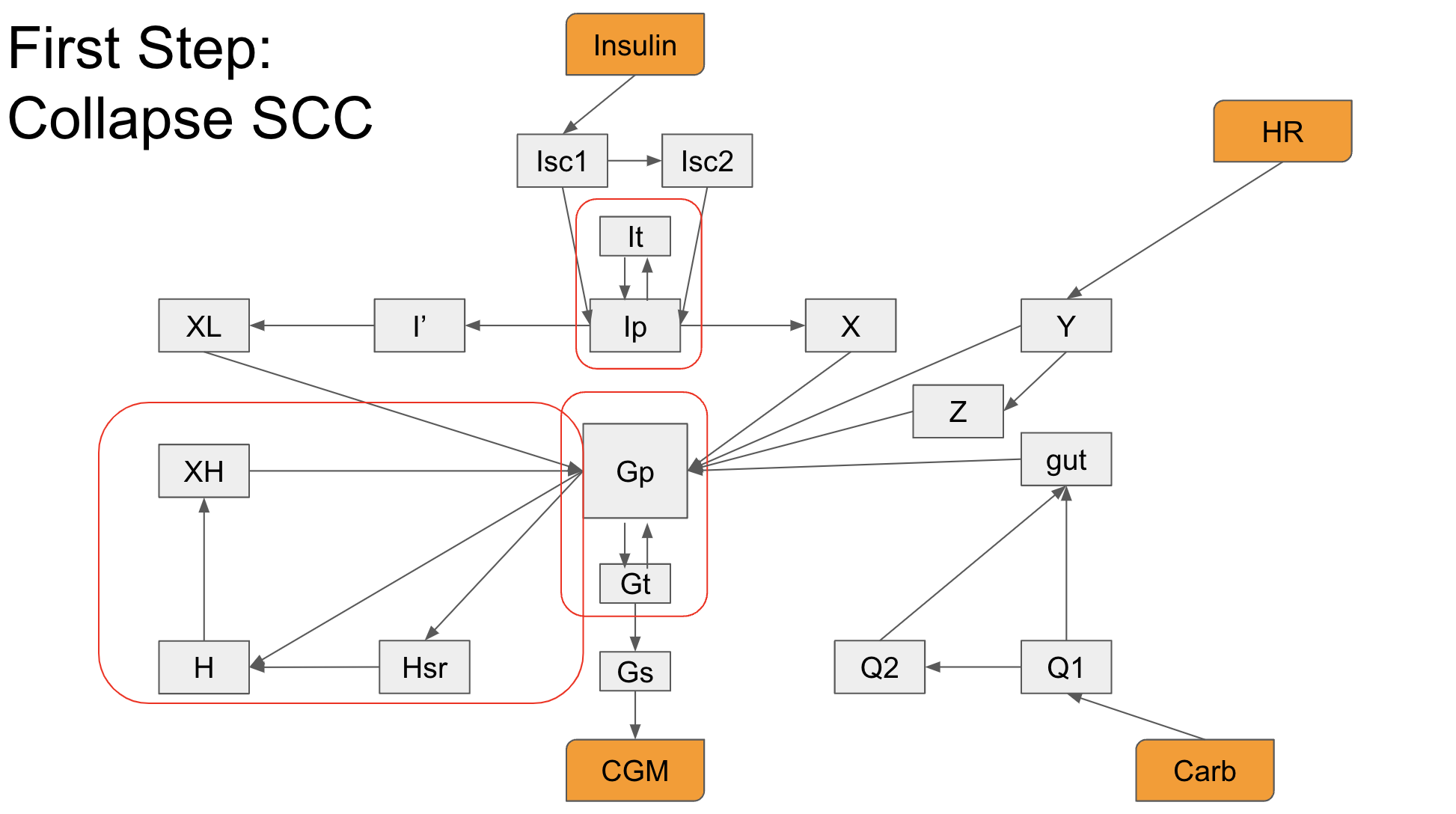}
         \caption{SCCs in the Starting Graph}
         \label{fig:showscc}
     \end{subfigure}
     \hfill
     \begin{subfigure}[b]{0.3\textwidth}
         \centering
         \includegraphics[width=\textwidth]{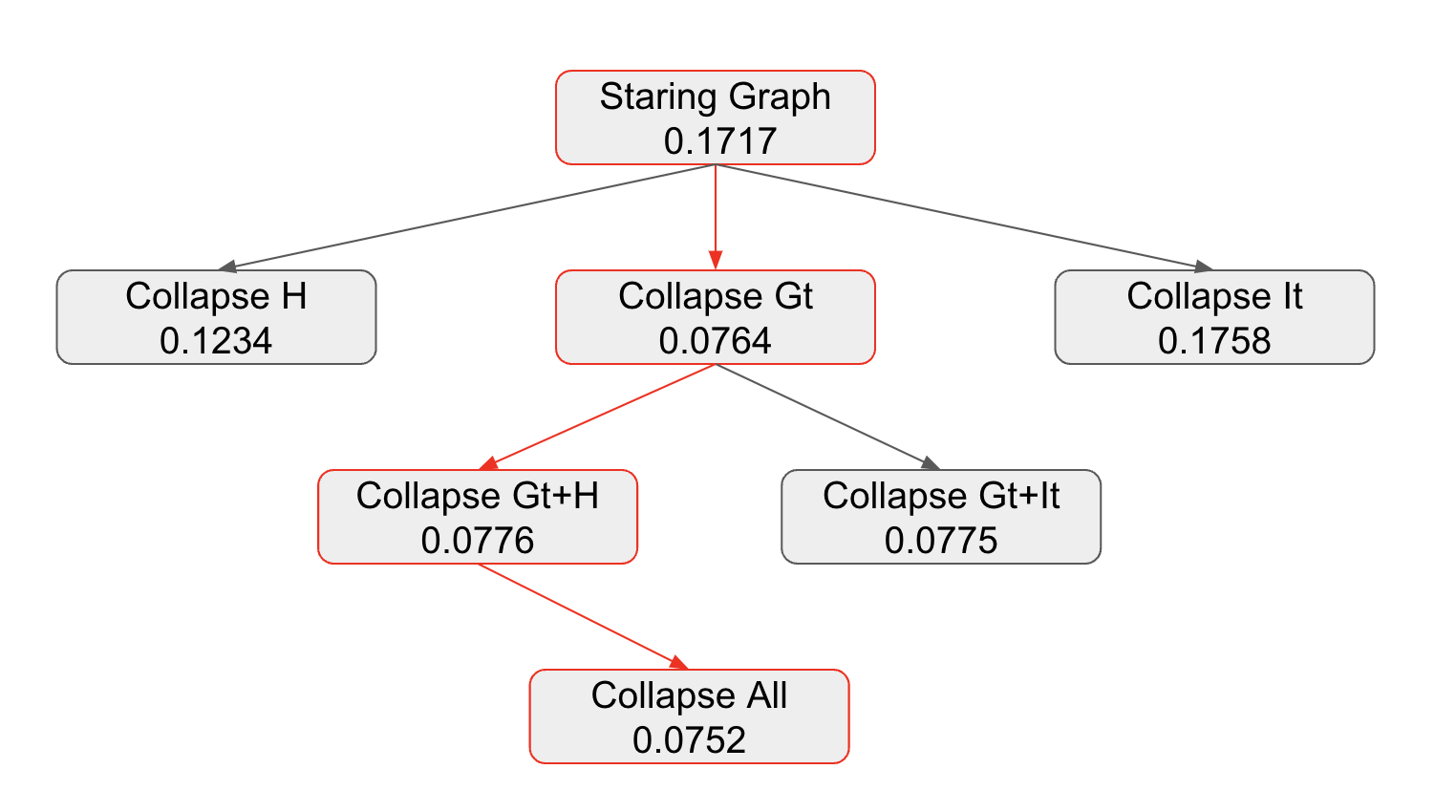}
         \caption{Reduction Tree of the SCCs}
         \label{fig:scctree}
     \end{subfigure}
     \hfill
     \begin{subfigure}[b]{0.3\textwidth}
         \centering
         \includegraphics[width=\textwidth]{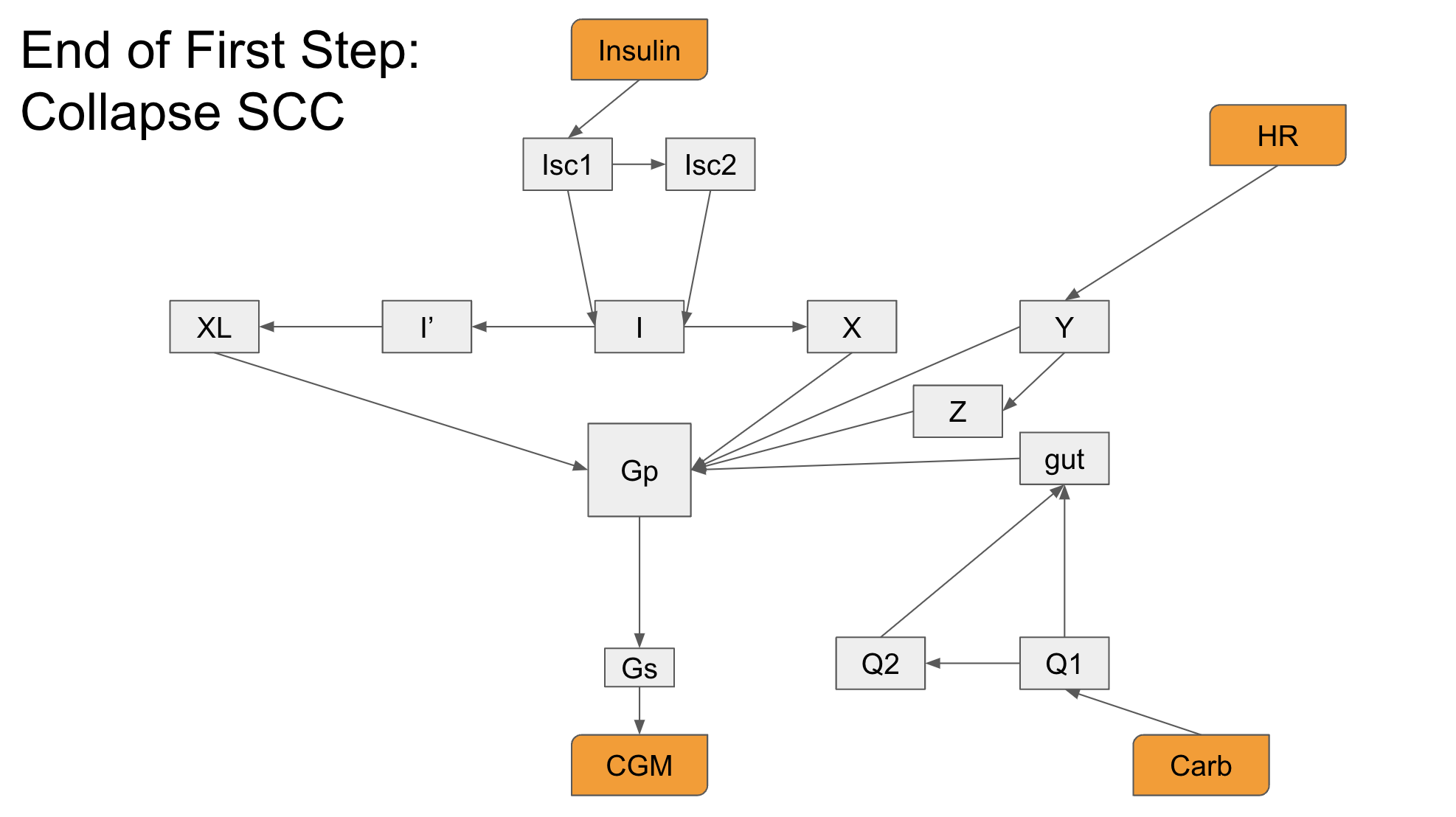}
         \caption{End Graph of the SCC Step}
         \label{fig:endofsccstep}
     \end{subfigure}
        \caption{Illustration of the Collapsing SCC Step of Our Heuristic}
        \label{fig:sccstep}
\end{figure}
    \item Step 3: For each group of non-overlapping paths with same source and destination nodes, try merge them by keeping only the path of greatest length and evaluate MNODE's performance with the resulting graph on the validation set. Adopt the change with the same criteria as in step 1.
    \item Step 4:  Repeat step 3 until no change satisfy both criteria. Figure Figure \ref{fig:combinepathstep} shows an illustrative summary of step 3 and step 4.
    \begin{figure}
     \centering
     \begin{subfigure}[b]{0.3\textwidth}
         \centering
         \includegraphics[width=\textwidth]{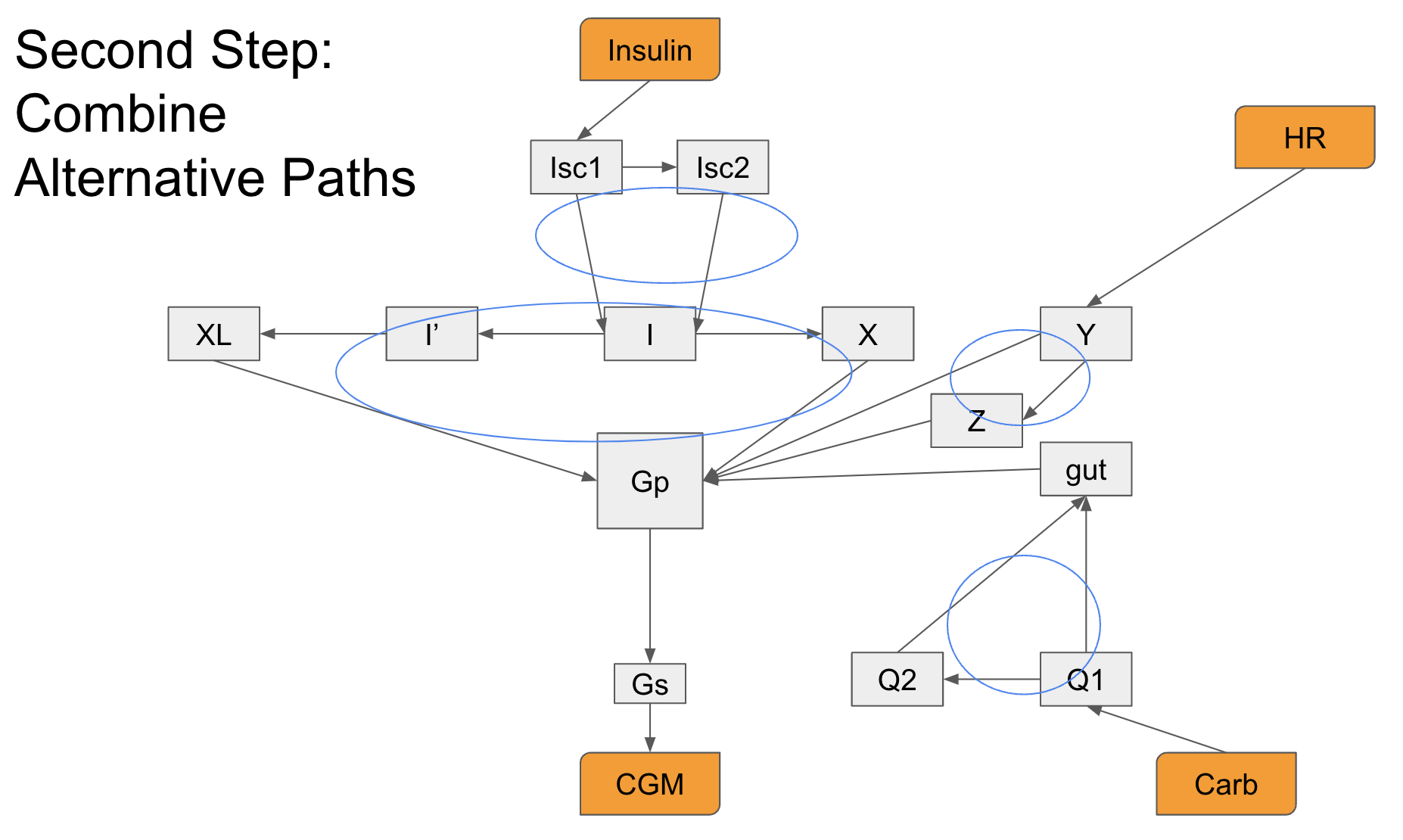}
         \caption{Non-overlapping Path Groups in the Starting Graph}
         \label{fig:showcombinepaths}
     \end{subfigure}
     \hfill
     \begin{subfigure}[b]{0.3\textwidth}
         \centering
         \includegraphics[width=\textwidth]{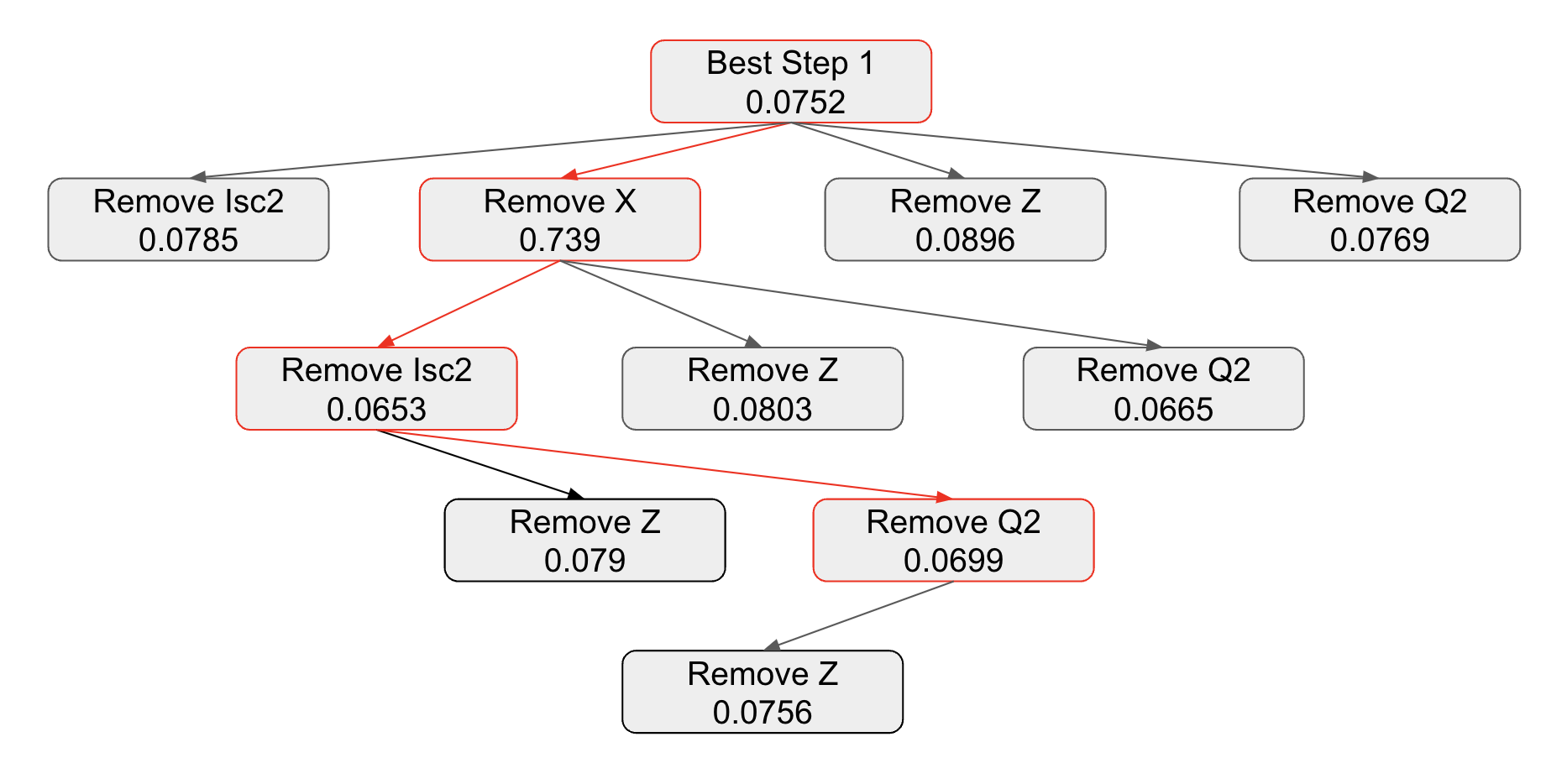}
         \caption{Reduction Tree of the Non-overlapping Paths}
         \label{fig:combinetree}
     \end{subfigure}
     \hfill
     \begin{subfigure}[b]{0.3\textwidth}
         \centering
         \includegraphics[width=\textwidth]{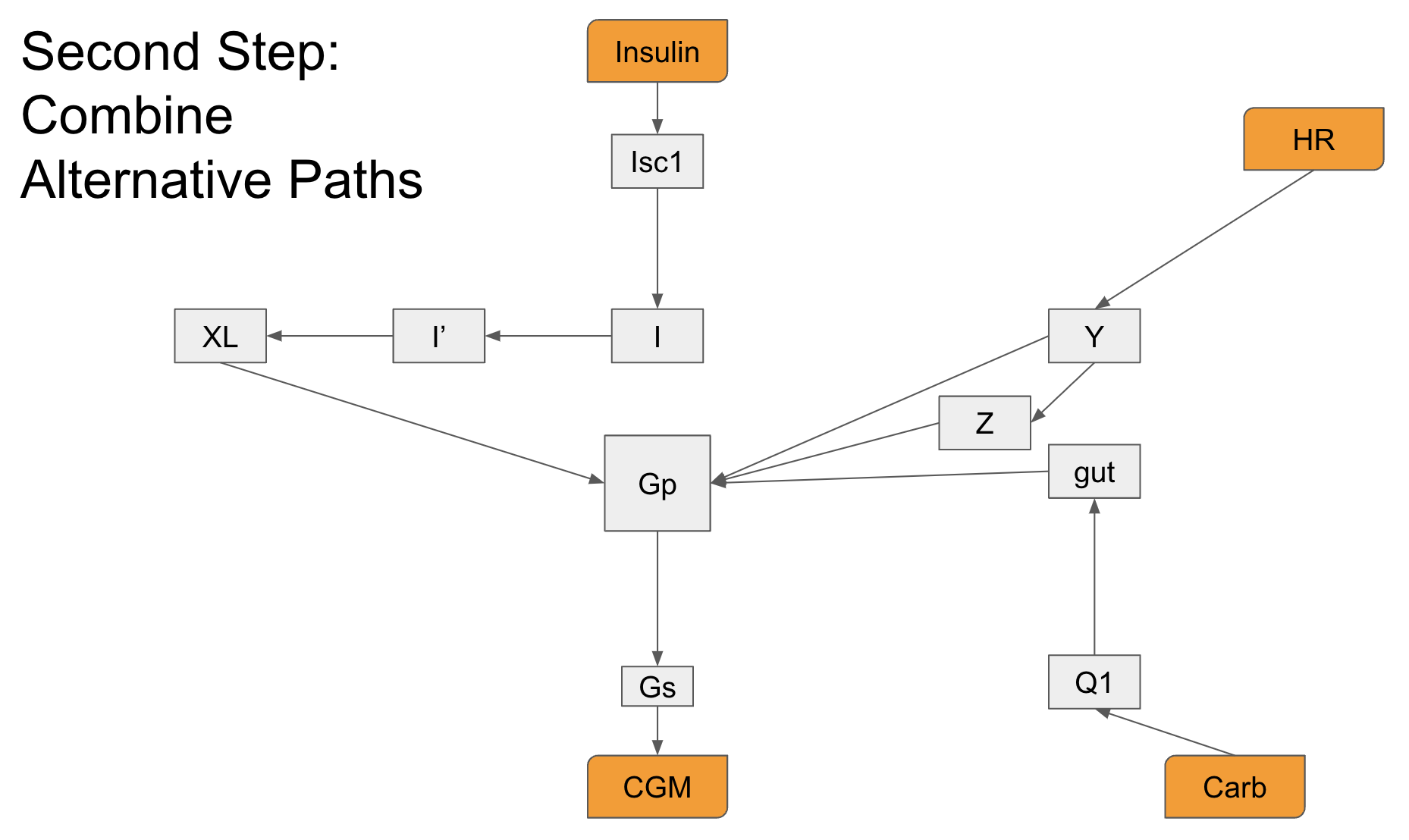}
         \caption{End Graph of the Path Combine Step}
         \label{fig:endofcombinepath}
     \end{subfigure}
        \caption{Illustration of the Combining Non-overlapping Paths Step of Our Heuristic}
        \label{fig:combinepathstep}
\end{figure}
    
   \item Step 5: For each path between an input node and the output node (at this point they should all be disjoint), try reducing its length by 1 via removing one intermediate node and evaluate MNODE's performance with the resulting graph on the validation set. Adopt the change with the same criteria as in step 1.
    \item Step 6:  Repeat step 5 until no change satisfy both criteria. Figure Figure \ref{fig:shortenpathstep} shows an illustrative summary of step 3 and step 4. Figure is the final graph we used in MNODE.
    \begin{figure}
     \centering
     \begin{subfigure}[b]{0.3\textwidth}
         \centering
         \includegraphics[width=\textwidth]{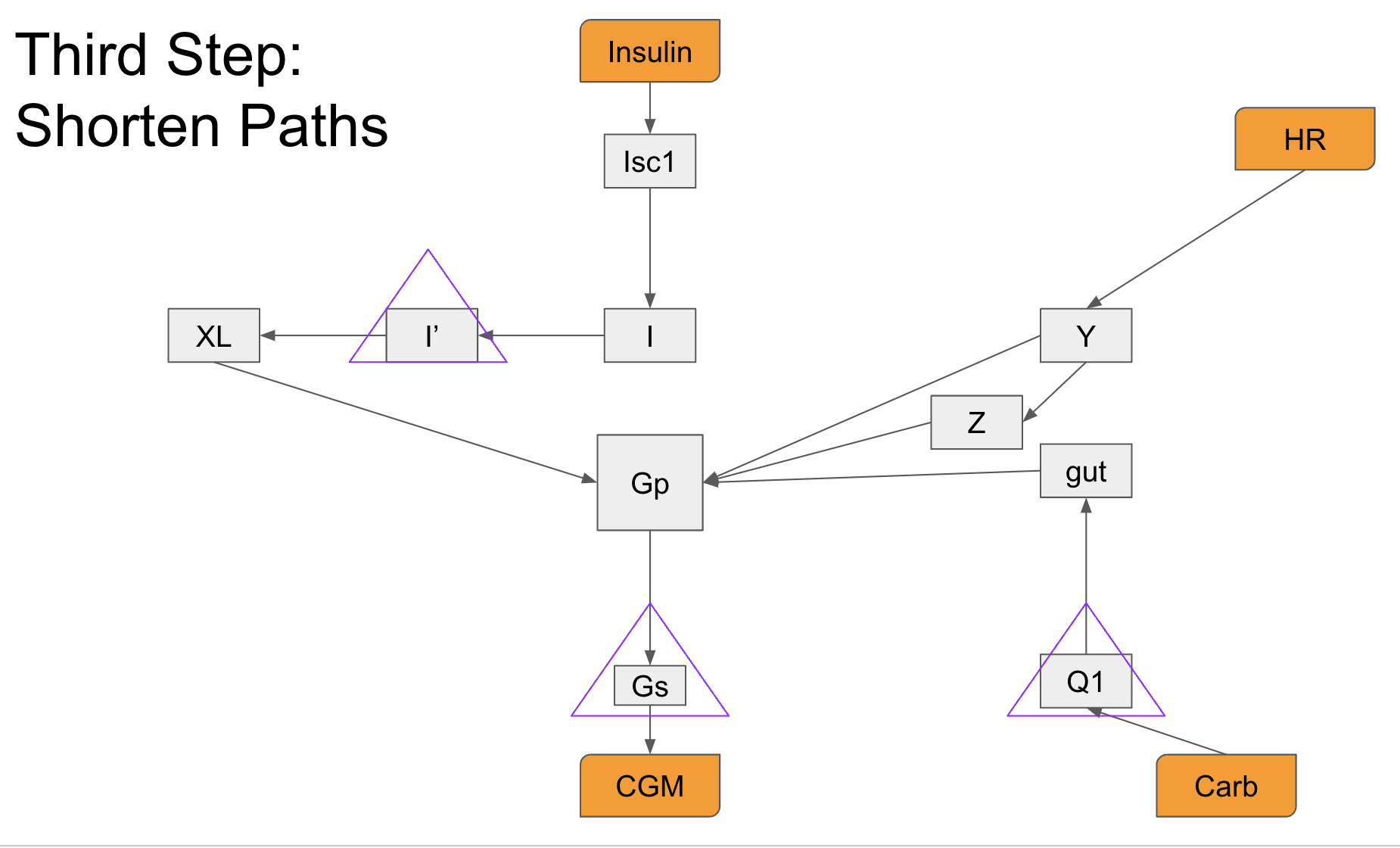}
         \caption{Paths to be Shortened in the Starting Graph}
         \label{fig:showshortenpaths}
     \end{subfigure}
     \hfill
     \begin{subfigure}[b]{0.3\textwidth}
         \centering
         \includegraphics[width=\textwidth]{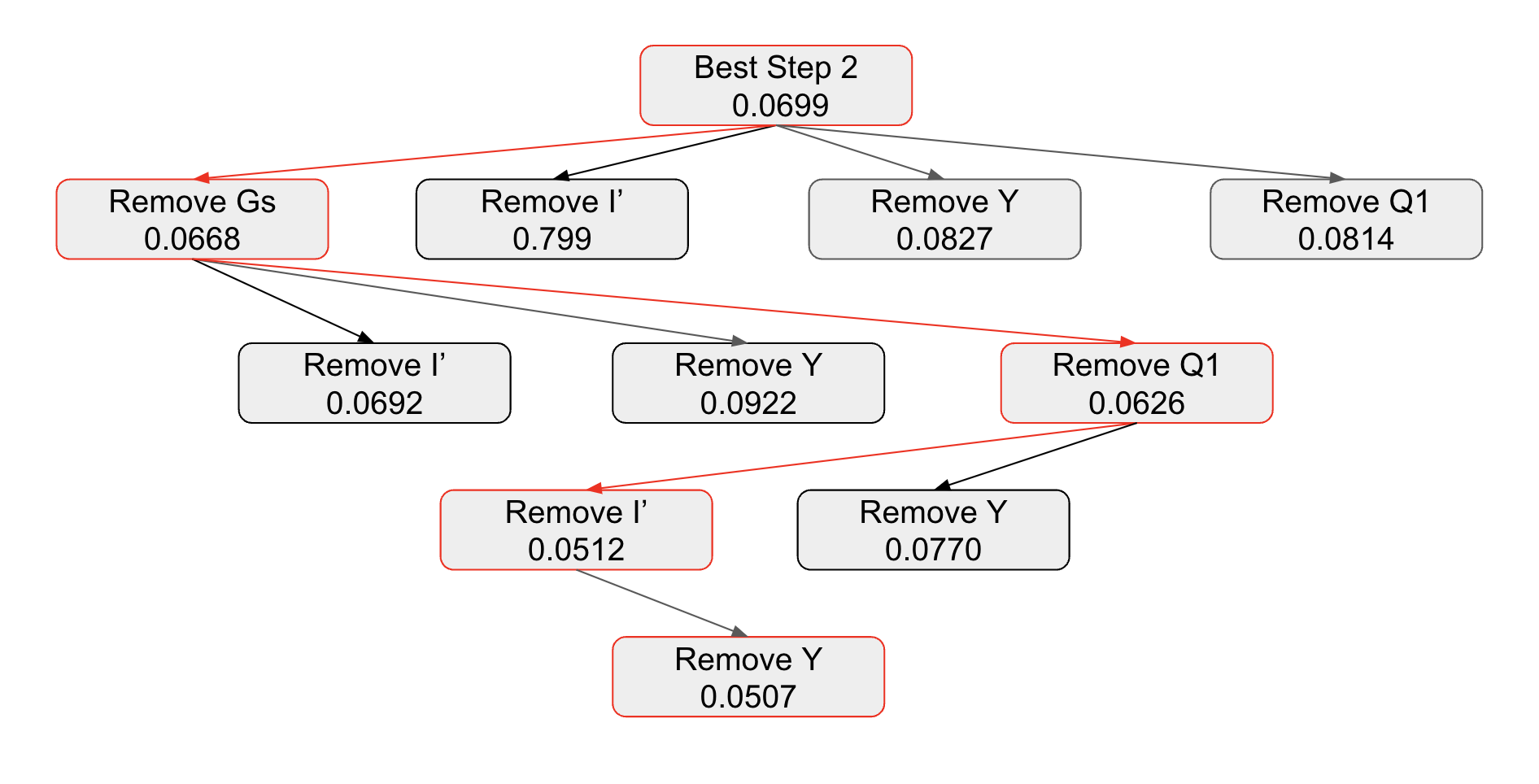}
         \caption{Reduction Tree of the Path Shortening Step}
         \label{fig:shorterntree}
     \end{subfigure}
     \hfill
     \begin{subfigure}[b]{0.3\textwidth}
         \centering
         \includegraphics[width=\textwidth]{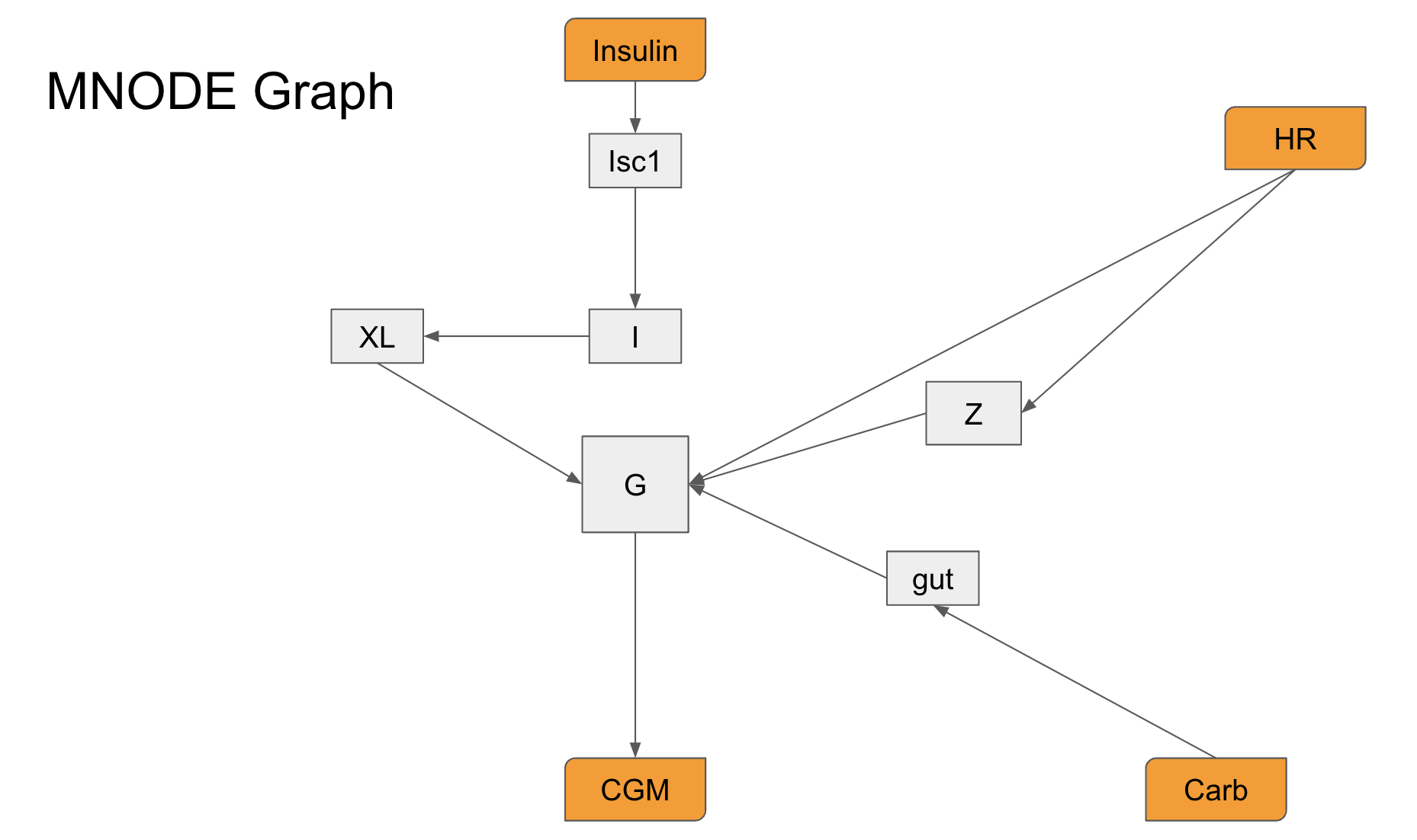}
         \caption{End Graph of the Path Shortening Step}
         \label{fig:endofshortenpath}
     \end{subfigure}
        \caption{Illustration of the Combining Non-overlapping Paths Step of Our Heuristic}
        \label{fig:shortenpathstep}
\end{figure} 
The final graph we use for MNODE is therefore Figure \ref{fig:finalgraph}, we showed it again with larger size for better readability.
\begin{figure}
    \centering
    \includegraphics[width=\linewidth/2]{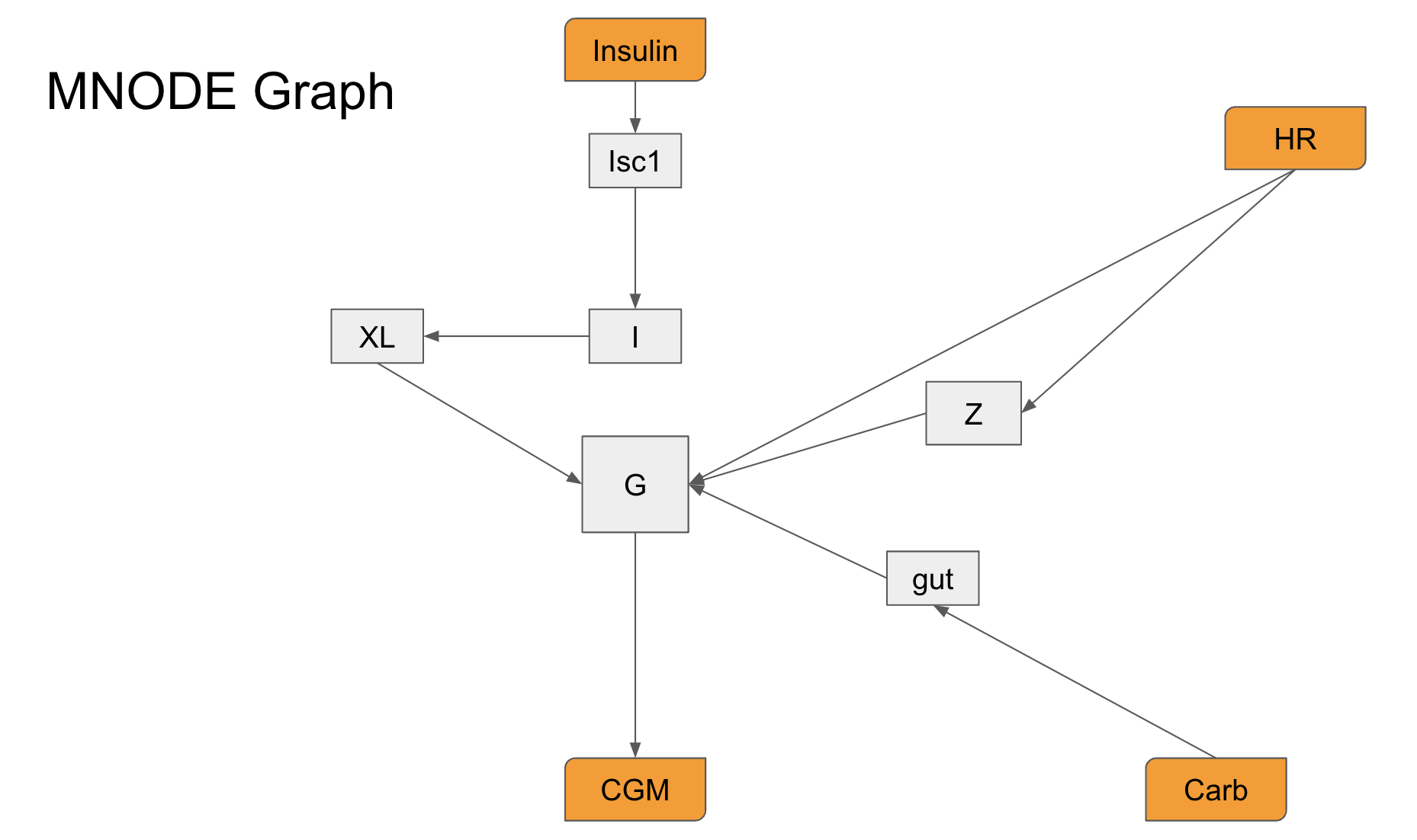}
    \caption{Final Graph used by MNODE}
    \label{fig:finalgraph}
\end{figure}
\end{enumerate}

\section{Detailed Data Preparation Procedure for T1DEXI}

\subsection{T1DEXI Exercise Instances}

Here we provide a step-by-step procedure of how we pre-processed the T1DEXI dataset described in Section \ref{sec:data}.

We select patients on open-loop pumps with age under 40 and body mass index (BMI) less than 30.  For each selected exercise instance in our dataset, we focus on the time window from 4 hours prior to the end of exercise, to 30 minutes after the end of exercise.  We anchor time zero as the end of exercise, so this time window (in minutes) is $[-240,30]$.  Since CGM measurements are taken in 5 minute increments, we divide this time window into 5 minute increments, and divide the {\em end time} of each interval by 5, so that we obtain 54 discrete time steps: $t = -47,-46,-45,\ldots, 4,5,6$, with $t = 0$ denoting the end of exercise.

    For each exercise instance, we use the following (54-dimensional) features derived from the T1DEXI data:
    \begin{enumerate}
    \item {\em CGM readings}.  CGM readings are taken every 5 minutes for all patients in T1DEXI.
    \item {\em Insulin injection}.  If the patient injected insulin at a given time in the T1DEXI dataset, we add that amount of insulin to the corresponding 5 minute interval to create a bolus insulin injection time series.  We add basal insulin to the bolus insulin time series to produce the (total) insulin time series used in our experiments.
    \item {\em Carbohydrate intake}.  Suppose the patient consumed carbohydrates at a given time in the T1DEXI dataset.  We assume a constant meal consumption rate of 45 grams per minute, and then compute the average consumption rate over each 5 minute interval to create a carbohydrate consumption time series.
    \item {\em Heart rate}.  We average the heart rate over each 5 minute interval to obtain a heart rate time series.
    \item {\em Step count}.  Similarly, we average step counts over each 5 minute interval to obtain a step count time series.
    \end{enumerate}

We only consider exercise instances for which:
\begin{enumerate}
    \item all 54 CGM readings were available;
    \item the patient logged carbohydrates 
    during the instance;
    \item the exercise duration was at least 30 minutes;
    % \delbz{planned}
    \item it was either the first or second exercise instance for that patient; and
    \item HR was consistently recorded every 10 seconds without any missingness throughout the 54 5-min intervals.
    % (see L224 in pre-processing code?)\revvbz{should be L203}
\end{enumerate}
After this process we end up with 143 exercise instances (13 patients with 1 exercise instance, and 65 patients with 2 exercise instances).

\begin{figure}[h]
    \centering
    \includegraphics[width=\linewidth]{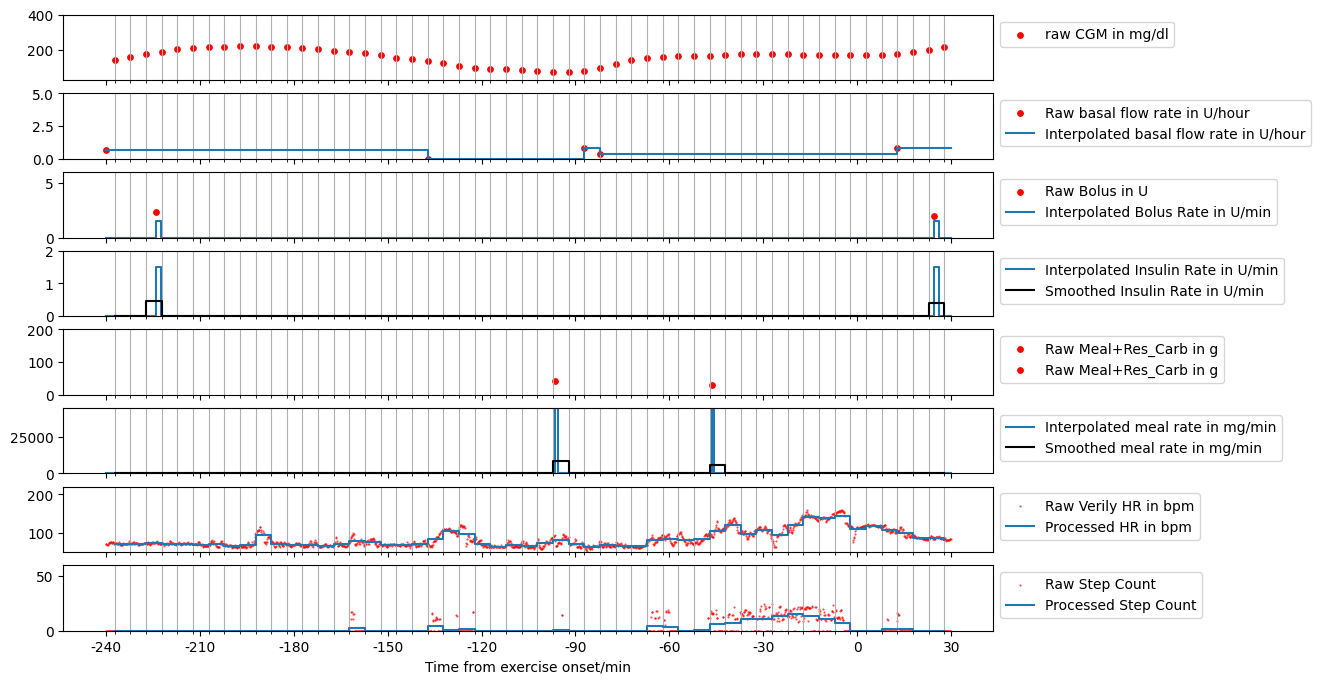}
    \caption{We show an example of the raw and processed data from a single patient window. Here, time $0$ represents the end of exercise, and we plot all features for the 4 hours prior to exercise termination, as well as the subsequent 30 minutes. In all plots, red dots indicate raw measurements, blue lines represent our clinically-informed representations and interpolations of the raw data, and the black curves (when relevant) represent subsequent mapping of the interpolated data to 5 minute intervals.}
    \label{fig:dataprocess}
\end{figure}

\Cref{fig:dataprocess} depicts the raw and processed data for a single exercise time-window for a single patient. The first row shows a complete timeseries of CGM, while the second row shows the recorded changes in basal insulin rates. While insulin pump data only includes the change points (red dots in the second plot), the devices function by executing the new constant rate (blue line) until a new basal rate is set (a subsequent red dot). The third plot shows bolus insulin recordings (red dots), which are sometimes thought of as being delivered instantaneously. However, insulin pumps in fact administer these doses by delivering insulin at a constant rate over a suitable short time window (typically $< 5$ min), such that the sum total of disbursed bolus (integral of the blue curve in the third plot) corresponds to the recorded dose (red dot). In the case of insulin pumps, both the basal and bolus insulins are fundamentally the same drug and are only distinguishable by their administration pattern; for this reason, we convert all insulin administration to $U/min$ and sum the basal and bolus, producing the black curve in the fourth figure.

The fifth plot shows the amount of carbohydrates ($g$) that were recorded by the participant (either as a meal or as part of the clinical study as ``rescue'' carbohydrates explicitly intended to mitigate hypoglycemia). We assume that meals are consumed at a constant $45 g/min$ and plot the corresponding carbohydrate consumption rate in the sixth plot (blue); we then align this impulse function with the 5 minute grid defined by CGM data (black). Finally, the last two figures show heart rate and step counts, respectively. In both cases, we average the raw data (red) over each 5 minute window, and perform piecewise-constant interpolation of these values (blue). Observe that step count and heart rate begin to rise at $t=-60$; the exercise lasts for approximately 1 hour, then the step counts and heart rates drop for $t>0$, once exercise is complete.

\subsection{Intervention Sets}

For each exercise time series, we first create an intervention set containing three copies of the original time series. And then we uniformly randomly (simulated by the numpy default random number generator with seed 2024) apply one of the four following intervention sets to it:
\begin{enumerate}
    \item Add 0/50/100 grams of carbohydrate to the three copies at the end of exercise (time step $t = 0$);
    \item Add 0/2.5/5 units of insulin to the three copies from the end of exercise onwards (time steps $t = 0, 1, \ldots, 6$);
    \item Add nothing/50 grams of carbohydrate/10 units of insulin to the three copies at the end of exercise (time step $t = 0$);
    \item Change the recorded heart rate in the last 7 time steps $t = 0, 1, \ldots, 6$ to typical medium intensity aerobic training (80,90,100,110,120,130,120)/ typical interval training (80,170,80,170,80,170,80)/typical high intensity resistance training (160,170,180,170,160,180,160).
\end{enumerate}
%To make sure the labels are not always the same for the same type of intervention, we then switch the 2nd and the 3rd copy with probability 0.5 (again simulated with the same numpy random number generator). 
For each exercise instance this leads to four time series (each consisting of 5 features and 54 time steps): the original observed time series, and three copies corresponding to the intervention set.

In addition, we compute the class label for each intervention set as the index of the intervention that leads to the highest mean glucose: 100 grams of carbohydrates for category 1; zero insulin for category 2; 50 grams of carbohydrate for category 3; and the highest heart rate level for category 4.
%inally we compute the corresponding class label of the intervention set and add it to the causal task solution data set.  

\section{Model Implementations and Hyperparameters}
\label{implementation-detail}
\paragraph{Set-up}For each model mentioned in the experiment section, here we offer a detailed description of the corresponding computational method and the hyperparameters used. Throughout this section we are given an exercise time series of 54 time steps (corresponding to 54 5 minute intervals that made up the time window starting from 4 hours prior to exercise termination, and ending at 30 minutes after exercise termination) and 5 features (corresponding to CGM reading, insulin, carbohydrate, heart rate, and step count, in that order). We denote the first feature (CGM reading) as $y$, and the other 4 features as $x$. We use a subscript to indicate discrete time steps and superscript to indicate feature indices. For example, $x_1^2$ is the 2nd feature of $x$ (carbohydrate) at discrete time step $t=1$. 

Our goal is to predict the CGM trace during the first 30 minutes following exercise completion corresponding to the output $y_{1:6}\in\mathbb{R}^6$ and therefore we set the number of prediction steps $q$ to be 6 for all models. We further split the given time series into historical context $p=(y,x)_{-47:-1}\in \mathbb{R}^{47\times 5}$, starting glucose $y_0\in \mathbb{R}$, inputs during exercise $x_{0:5}\in \mathbb{R}^{6\times 4}$ (six inputs that are recorded 1 time step ahead of the expected outputs).  We use $\hat{y}\in\mathbb{R}^{6}$ to indicate the CGM trace predicted by models, $s$ for modeled states and $z$ for latent states. $h,c$ for the final hidden state and cell state of the LSTM initial condition learner. For ease of computation and without loss of generality, we set the $\Delta t$ term in forward-Euler style discretization to be $1$ for all relevant models, and thus we omit it in the equations.
\paragraph{Learning Rate, Initialization and Optimizer} For all experiments, we use the Adam optimizer \citep{kingma2014adam} to perform gradient descent.  

Learning rates and weight initialization are set with model stability as a consideration.  In particular, in all experiments using T1DEXI data, we set the default learning rate to be $2\times 10^{-3}$ for all models except UVA, for which we use a larger learning rate of $10^{-1}$.  For all models except UVA, we initialize model weights with the PyTorch default setting if they are part of a pre-defined PyTorch model class or standard normal if they are custom weights.
For UVA we initialize model weights from a normal distribution with mean zero and variance $1/400$.  UVA is treated differently because ODE-based mechanistic models are notoriously highly sensitive to parameter initialization, and it is both important to initialize the parameters to be small in magnitude, and to make sure they do not get too close to zero during training.

For synthetic experiments, we set the learning rate of the purely mechanistic model to $5 \times 10^{-1}$; to $1 \times 10^{-2}$ for LP and LPSC; and to $2 \times 10^{-3}$ for the remainder of the models.

% For synthetic experiments, 
% \revvbz{except that we set the learning rate for UVA mechanistic model to be $1\cdot 10^{-1}$ for stability considerations. This is because ODE-based mechanistic models are highly sensitive to parameter initialization, and it is both important to initialize the parameters to be small in magnitude and to make sure they do not get too close to 0 during training. We deal with this challenge by initializing the model parameters from a normal distribution with mean 0 and variance $1/400$ and adopting a larger learning rate at the same time. We initialize all other models weights with PyTorch default setting if they are part of a pre-defined PyTorch model class or standard normal if they are custom weights.}  In synthetic experiments, \revvbz{while stability is less of a concern due to the simple structure of the system, we still set the learning rate for purely mechanistic model to be $5\cdot 10^{-1}$ and the learning rate for LP and LPSC to be $1\cdot 10^{-2}$ for faster and more stable convergence, and the rest of the models still use a learning rate of $2\cdot 10^{-3}$. We keep weight initialization setting the same as the TIDEXI experiments.}

\paragraph{Training Epochs}

Unless otherwise specified in the model description (LPSC), in T1DEXI experiments, we train for 100 epochs and pick the epoch with best validation loss. In synthetic experiments, we train for 50 epochs and pick the epoch with best validation loss. 

\paragraph{Dropout} For the hybrid models, we use 0 dropout rate. This is because we are already doing regularization via early stopping, and our trial runs indicate that there is no need for additional dropout. This is not true for black-box models, for which we still tune the dropout rate.

\paragraph{Hyperparameter Search} We use grid search to tune hyperparameters. When choosing the grid, we restrict the search space to areas where the models have less than 20000 parameters and we also try to limit the number of grid points to around 10. This is to make sure the computational cost of the experiments is capped at a reasonable level for small data sets and individual users. The grid used for each model in each experiment will be provided below together with model descriptions.

\subsection{UVA}
\label{sec:UVA}
UVA is our baseline mechanisitc model, whose description is given in Appendix A. The computation equation for the UVA model is given in Algorithm \ref{alg:uva}.
\begin{algorithm}[th]
   \caption{UVA Padova Model}
   \label{alg:uva}
\begin{algorithmic}
   \STATE {\bfseries Input:} number of prediction steps $q=6$, historical context $p$, starting glucose $y_0$, exogenous inputs $x_{0:5}$, the original UVA mechanistic ODEs $m_{\text{UVA}}$, $\Delta t=1$
   \STATE $h,c=\text{LSTM}(p)$
   \STATE $h^1=y_0$
   \STATE $s_0=h$\\
   \FOR{$i=0:q-1$}
   \STATE $s_{i+1}=s_i+\Delta t\cdot m_{\text{UVA}}(s_i,x_i;\beta)$
   \STATE $\hat{y}_{i+1}=s_{t+1}^1$  
   \ENDFOR
   \STATE {\bfseries Output:} $\hat{y}_{1:q}$
\end{algorithmic}
\end{algorithm}

\paragraph{T1DEXI Experiments} The LSTM network has 2 layers and 21 hidden dimensions (same as the UVA/Padova model), and we set the first state of the estimated initial condition (represented by hidden state $h$) to be the true initial value of CGM $y_0$ to keep consistency with the assumption that $s^1$ represents glucose. The UVA/Padova Simulator has 53 trainable parameters $\beta\in\mathbb{R}^{53}$. We do not tune hyperparameters for the UVA model.

\paragraph{Synthetic Experiments} In the synthetic setting, we assume to know the ground truth model. And therefore we remove the LSTM initial condition encoder and instead set $s_0=y_0$ since in the synthetic setting, there is only one state. Same as T1DEXI, we do not tune hyperparameters.

\subsection{Reduced UVA Latent Parameter Learning}
The Reduced UVA Latent Parameter Learning model uses a lower-fidelity, reduced version of UVA/Padova defined in Appendix \ref{app:ruva}, and applies the idea of latent parameter learning to it. The computation equations are given in Algorithm \ref{alg:dtd}
\begin{algorithm}[th]
   \caption{Reduced UVA Latent Parameter Learning Model}
   \label{alg:dtd}
\begin{algorithmic}
   \STATE {\bfseries Input:} number of prediction steps $q$, historical context $p$, starting glucose $y_0$, exogenous inputs $x$, the reduced UVA mechanistic ODEs $m_{\text{RUVA}}$, $\Delta t =1$
   \STATE $h,c=\text{LSTM}(p)$
   \STATE $s_0=\text{concatenate}(y0,\text{MLP1}(h))$\\
   \STATE $z_0=c$
   \FOR{$i=0:q-1$}
   \STATE $s_{i+1}=s_i+\Delta t \cdot m_{\text{RUVA}}(s_i,x_i;\beta_i=\text{MLP2}(z_i))$
   \STATE $z_{i+1}=Az_i+Bx_i^{3\sim 4}$
   \STATE $\hat{y}_{i+1}=s_{t+1}^1$  
   \ENDFOR
   \STATE {\bfseries Output:} $\hat{y}_{1:q}$
\end{algorithmic}
\end{algorithm}

\paragraph{T1DEXI Experiments} In our implementation, the latent dynamics of $z$ only depends on itself and inputs that are not used by the UVA S2013 model (the 3rd and 4th feature, which correspond to heart rate and step count). We made this choice to keep consistency with the implementation in \citet{miller2020learning}.
The LSTM has 2 layers and $d$ hidden dimension (since this time the cell states are used to initialize latent state $z$, whose dimension needs to be tuned) and MLP2 serves to map $d$ dimensional $h_0$ to a 8-dimensional initial condition vector. All MLPs have $n$ hidden layers and $m$ hidden units with dropout $0$ and activation ReLu, we tune these hyperpameters with grid search on the following grid:
\[n=\{2,3,4\}\times m=\{16,32,48\}\times d=\{8,12,16\}\]

\paragraph{Synthetic Experiments} Similar to the purely mechanistic model, in the synthetic setting we directly set $s_0=y_0$ and therefore do not use $h$ or MLP1. We still need the 2-layer LSTM to generate $z_0$. The search grid for synthetic experiments is:
\[n=\{2,3\}\times m=\{16,32\}\times d=\{2,4\}\]

\subsection{Reduced UVA Latent Parameter and State Closure Learning}
The Reduced UVA Latent Parameter and State Closure Learning model uses the same reduced UVA/Padova defined in Appendix \ref{app:ruva}, and applies both the idea of latent parameter learning and state closure learning to it. The computation equations are given in Algorithm \ref{alg:dtdc}
\begin{algorithm}[th]
   \caption{Reduced UVA Latent Parameter and State Closure Learning Model}
   \label{alg:dtdc}
\begin{algorithmic}
   \STATE {\bfseries Input:} number of prediction steps $q=6$, historical context $p$, starting glucose $y_0$, exogenous inputs $x_{0:5}$, the reduced UVA mechanistic ODEs $m_{\text{RUVA}}$ the adjacency matrices (as defined in Section \ref{sec:methods}) of reduced UVA: $A_s,A_x$, closure switch constant $w$, $\Delta t=1$ 
   \STATE $h,c=\text{LSTM}(p)$
   \STATE $s_0=\text{concatenate}(y_0,\text{MLP1}(h))$
   \STATE $z_0=c$
   \FOR{$i=0:q-1$}
   \STATE $s_{i+1}=s_i+\Delta t \cdot m_{\text{RUVA}}(s_i,x_i;\beta_i=\text{MLP2}(z_i))+w\cdot \text{MLPs}(s_i,x_i;A_s,A_x)$
   \STATE $z_{i+1}=Az_i+Bx_i^{3\sim 4}$
   \STATE $\hat{y}_{i+1}=s_{t+1}^1$  
   \ENDFOR
   \STATE {\bfseries Output:} $\hat{y}_{1:q}$
\end{algorithmic}
\end{algorithm}

\paragraph{T1DEXI Experiments} The LSTM has 2 layers and $d$ hidden dimension (since this time the cell states are used to initialize latent state $z$, whose dimension needs to be tuned). MLP1 and MLP2 have $n$ hidden layers and $m$ hidden units with dropout $0$ and activation ReLu, while MLPs that form the closure learning network have $2$ hidden layers and $m$ hidden units with dropout $0$ and activation ReLu. We tune these hyperpameters with grid search on the following grid:
\[n=\{2,3\}\times m=\{16,32\}\times d=\{8,16\}\]
Note that our search space is smaller as the introduction of extra MLPs significantly increases the total number of parameters.

\paragraph{Synthetic Experiments} We again set $s_0=y_0$ and do not use $h$ or MLP1. The hyperparameter grid is:
\[n=\{2,3\}\times m=\{16,32\}\times d=\{8,16\}\]

\paragraph{Special Training Routine}The training routine of models adopting state closure is more complicated. We first set $w=0$ and just train the reduced UVA latent parameter model and the initial condition learner LSTM for 100(T1DEXI)/50(Synthetic) epochs (the closure part is masked out by $w$). Then we freeze the weights of the reduced UVA latent parameter model and the LSTM, set $w=1$ and train the closure neural network on the residuals for 50 more epochs. This trick makes sure the closure network is truly learning the residuals as intended and not subsuming the mechanistic model.
\subsection{Mechanistic Neural ODE}
The MNODE model no longer relies on the functional forms of UVA models, and therefore does not require any ODE equations $m$. Instead, it uses the adjacency matrices of a reduced version of the UVA/Padova graph. We obtain this reduced graph with the reduction heuristic described in Appendix \ref{app:graphred}. We provide its computation equations in Algorithm \ref{alg:mnode}.
\begin{algorithm}[th]
   \caption{Mechanistic Neural ODE Model}
   \label{alg:mnode}
\begin{algorithmic}
   \STATE {\bfseries Input:} number of prediction steps $q=6$, historical context $p$,starting glucose $y_0$,  exogenous inputs $x_{0:5}$, the adjacency matrices (as defined in Section \ref{sec:methods}) of the reduced UVA graph $A_s,A_x$, $\Delta t=1$
   \STATE $h,c=\text{LSTM}(p)$
   \STATE $h^1=y_0$
   \STATE $s_0=h$
   \FOR{$i=0:q-1$}
   \STATE $s_{i+1}=s_i+\Delta t \cdot \text{MLPs}(s_i,x_i;A_s,A_x)$
   \STATE $\hat{y}_{i+1}=s_{t+1}^1$  
   \ENDFOR
   \STATE {\bfseries Output:} $\hat{y}_{1:q}$
\end{algorithmic}
\end{algorithm}
\paragraph{T1DEXI Experiments} The LSTM has 2 layers and $5$ hidden dimension (this corresponds to the number of states in the reduced graph) and for the same reason we set the 1st features of the initial condition state vector to be $y_0$. All MLPs have $n$ hidden layers and $m$ hidden units with dropout $0$ and activation ReLu, we tune these hyperpameters with grid search on the following grid:
\[n=\{2,3\}\times m=\{16,24,32\}.\]
\paragraph{Synthetic Experiments} We set $s_0=y_0$ and do not use $h$ or the LSTM (note that our code implementation still has the LSTM component for the sake of consistency but its output is actually never used by MNODE.) The search grid  is:
\[n=\{2,3\}\times m=\{16,32\}.\]

\subsection{BNODE}
For BNODE and the subsequent black-box models, we point the reader to implementations referenced in the associated citations in the main paper.  Here we describe our implementation.
\begin{algorithm}[th]
   \caption{Black-box Neural ODE Model}
   \label{alg:bnode}
\begin{algorithmic}
   \STATE {\bfseries Input:} number of prediction steps $q=6$, historical context $p$, starting glucose $y_0$, exogenous inputs $x_{0:5}$, $\Delta t=1$
   \STATE $h,c=\text{LSTM}(p)$
   \STATE $h^1=y_0$
   \STATE $s_0=h$
   \FOR{$i=0:q-1$}
   \STATE $s_{i+1}=s_i+\Delta t \cdot \text{MLPs}(s_i,x_i)$
   \STATE $\hat{y}_{i+1}=s_{t+1}^1$  
   \ENDFOR
   \STATE {\bfseries Output:} $\hat{y}_{1:q}$
\end{algorithmic}
\end{algorithm}

\paragraph{T1DEXI Experiments} The LSTM has 2 layers and $d$ hidden dimension. Note that here the hidden dimension of LSTM also determines the state dimension of the neural ODE, which is a tunable hyperparameter. All MLPs have $n$ hidden layers and $m$ hidden units with dropout $a$ and activation ReLU, we tune these hyperpameters with grid search on the following grid:
\[d=\{4,5,6\}\times n=\{2,3\}\times m=\{32,48,60\}\times a=\{0,0.1,0.2\}.\]
\paragraph{Synthetic Experiments} We keep all the settings the same (note we can no longer assume there is only 1 state as in the black-box regime we have zero domain knowledge) as the T1DEXI experiments except that the search grid for hyperparameters is now:
\[d=\{2,3\}\times n=\{2,3\}\times m=\{64\}\times a=\{0,0.2\}.\]

\subsection{TCN}
\begin{algorithm}[th]
   \caption{Temporal Convolutional Network Model}
   \label{alg:tcn}
\begin{algorithmic}
   \STATE {\bfseries Input:} number of prediction steps $q=6$, historical context $p$, starting glucose $y_0$,  exogenous inputs $x_{0:5}$
   \STATE $\tilde{x} = \mathbf{0}\in\mathbb{R}^q$
   \STATE $\tilde{x}_0=y_0$
   \STATE $x'=\text{concatenate}(\tilde{x},x,\text{dim}=-1)$
   \STATE $seq_{in}=\text{concatenate}(p,x',\text{dim}=0)$
   \STATE $seq_{out}=\text{TCN}(seq_in)$
   \STATE $\hat{y}=\text{Linear}(seq_{out})$  
   \STATE {\bfseries Output:} $\hat{y}$
\end{algorithmic}
\end{algorithm}
\paragraph{T1DEXI Experiments} The TCN model is taken directly from the code repository posted on \url{https://github.com/locuslab/TCN/blob/master/TCN/tcn.py}, with input size set to 5, number of channels set to a list of $n$ copies of $m$, kernel size set to $l$ and dropout set to $a$. We tune these hyperpameters with grid search on the following grid:
\[n=\{2,3\}\times m=\{16,24,32\}\times l=\{2,3,4\} \times a=\{0,0.1,0.2\}.\]

\paragraph{Synthetic Experiments} We keep all the settings the same as the T1DEXI experiments except that the input size is set to 3 and the search grid for hyperparameters is now:
\[n=\{2,3\}\times m=\{16,32\}\times l=\{2,3,4\} \times a=\{0,0.2\}.\]

\subsection{LSTM}
\begin{algorithm}[th]
   \caption{Long Short Term Memory Model}
   \label{alg:lstm}
\begin{algorithmic}
    \STATE {\bfseries Input:} number of prediction steps $q=6$, historical context $p$, starting glucose $y_0$,  exogenous inputs $x_{0:5}$
   \STATE $h,c=\text{Encoder LSTM}(p)$
   \STATE Set initial hidden state and cell state of Decoder LSTM to $h,c$ respectively
   \STATE $seq_{out}, h_q,c_q=\text{Decoder LSTM}(x_{0:5})$
   \STATE $\hat{y}=\text{Linear}(seq_{out})$
   \STATE {\bfseries Output:} $\hat{y}$
\end{algorithmic}
\end{algorithm}
\paragraph{T1DEXI Experiments} Both Encoder and Decoder LSTM have $n$ layers and $d$ hidden states with dropout set to $a$. We tune these hyperpameters with grid search on the following grid:
\[n=\{2,3.4\}\times m=\{8,12,16\}\times a=\{0,0.1,0.2\}.\]
\paragraph{Synthetic Experiments}
\[n=\{2,3\}\times m=\{8,16\}\times a=\{0,0.2\}.\]
\subsection{Transformer}
\begin{algorithm}[th]
   \caption{Transformer Model}
   \label{alg:trans}
\begin{algorithmic}
   \STATE {\bfseries Input:} number of prediction steps $q=6$, historical context $p$, starting glucose $y_0$,  exogenous inputs $x_{0:q-1}$, true output $y_{1:q-1}$ (needed during training)
   \STATE $\tilde{x} = \mathbf{0}\in\mathbb{R}^q$
   \STATE $\tilde{x}_0=y_0$
   \STATE $x'=\text{concatenate}(\tilde{x},x,\text{dim}=-1)$ 
   \STATE $\text{encoder\_in}=\text{concatenate}(p,x',\text{dim}=0)$ (concatenating all inputs to form a masked context)
\IF{Model in Training Mode}
    \STATE $\text{decoder\_in} = \text{concatenate}(y_0, y_{1:q-1})$ (expected output shifted to the right)
    \STATE $\text{decode\_out} = \text{Transformer}(\text{encoder\_in}, \text{decoder\_in}, \text{decoder\_causal\_mask})$
\ENDIF
\IF{Model in Evaluation Mode}
    \STATE $\text{decoder\_in} = \text{concatenate}(y_0, \mathbf{0} \in \mathbb{R}^{q-1})$
    \FOR{$i = 1 : q-1$}
        \STATE $\text{decode\_out} = \text{Transformer}(\text{encoder\_in}, \text{decoder\_in})$
        \STATE $\text{decoder\_in}_{i+1} = \text{decode\_out}_{i}$
    \ENDFOR
    \STATE $\text{decode\_out} = \text{Transformer}(\text{encoder\_in}, \text{decoder\_in})$
\ENDIF
   \STATE $y=\text{Linear}(\text{decode\_out})$  
   \STATE {\bfseries Output:} $y$
\end{algorithmic}
\end{algorithm}
\paragraph{T1DEXI Experiments}We use the transformer model provided by the pytorch nn class, and its hyperparameters are set as follows: d\_model set to $d$, number of encoder layers set to $l_1$, number of decoder layers set to $l_2$, the dim\_feedforward is set to $m$ and dropout is set to $a$. We tune the hyperparameters with the following grid:
\[d=\{4,8\}\times l_1=\{2,3\}\times l_2=\{2,3\}\times m=\{32,64\}\times a=\{0,0.1\}\]
\paragraph{Synthetic Experiments} We keep all experiment settings the same except setting $q=10$ and the hyperparameter search grid to be:
\[d=\{4,8\}\times l_1=\{2\}\times l_2=\{2\}\times m=\{32,64\}\times a=\{0,0.2\}\]

\subsection{S4D}
\begin{algorithm}[th]
   \caption{S4 Diagonal Model}
   \label{alg:s4d}
\begin{algorithmic}
   \STATE {\bfseries Input:} number of prediction steps $q=6$, historical context $p$, starting glucose $y_0$,  exogenous inputs $x_{0:q-1}$
   \STATE $\tilde{x} = \mathbf{0}\in\mathbb{R}^q$
   \STATE $\tilde{x}_0=y_0$
   \STATE $x'=\text{concatenate}(\tilde{x},x_{0:5},\text{dim}=-1)$
   \STATE $seq_{in}=\text{concatenate}(p,x',\text{dim}=0)$
   \STATE $seq_{in}=\text{Linear}(Seq_{in})$
   \STATE $seq_{in}=\text{Transpose}(Seq_{in},1,2)$
   \STATE $seq_{out}=\text{S4D}(seq_in)$
   \STATE $seq_{out}=\text{Transpose}(Seq_{out},1,2)$
   \STATE $\hat{y}=\text{Linear}(seq_{out})_{-q:}$  
   \STATE {\bfseries Output:} $\hat{y}$
\end{algorithmic}
\end{algorithm}
\paragraph{T1DEXI Experiments} We take the S4D model directly from the following github repository \url{https://github.com/thjashin/multires-conv/blob/main/layers/s4d.py}, and its hyperparameters are set as: d\_model set to $d$, d\_state set to $m$, dropout set to $a$.We tune the hyperparameters with the following grid:
\[ d=\{4,6,8\}\times \{m=\{32,64\}\times a=\{0,0.1,0.2\}\]
\paragraph{Synthetic Experiments} We set $q=10$ and the hyperparameter search grid to be:
\[ d=\{3,4,5\}\times \{m=\{32,64\}\times a=\{0,0.2\}\]

\section{Repeated Nested Cross Validation}
We use the following algorithm to evaluate both predictive and causal generalization errors of our models.
\begin{algorithm}[H]
   \caption{Repeated Nested Cross Validation}
   \label{alg:rncv}
\begin{algorithmic}
   \STATE {\bfseries Input:} Data $D$, Model $M$, Hybrid Loss Function $l_h$, Alpha for Hybrid Loss $\alpha$,  List of Hyperparameter Settings $\Lambda$,\\ 
   Number of Repeats $R=3$, Number of Outer Fold $N=6$, Number of Inner Fold $M=4$, Random Seed $s=2024$
   \STATE Generate $R$ different permutations of $[1,\dots,\text{length}(D)]$: $P_1,\dots,P_R$ with numpy default random number generator and seed $s$
   \STATE Initialize Error Lists $e_{\text{pred}},e_{\text{causal}}$ 
   \FOR{$r=1:R$}
   \STATE Permute $D$ with permutation $P_r$, save the resulting data as $D_r$
   \STATE Split $D_r$ into $N$ folds $D_r^1,\dots D_r^N$
   \FOR{$i=1:N$}
   \STATE Form the outer training set as $D_{\text{tout}}=D_r\setminus D_r^i$, and set the test set as $D_{\text{test}}=D_r^i$
   \STATE Split $D_{\text{tout}}$ into M folds $D_{\text{tout}}^1,\dots,D_{\text{tout}}^M $
   \FOR{$j=1:M$}
   \STATE Form the training set as $D_{\text{train}}=D_{\text{tout}}\setminus D_{\text{tout}}^j$ 
   \STATE Form the validation set as $D_{\text{val}}=D_{\text{tout}}^j$
   \IF{$j<M$}
   \FOR{each $\lambda\in\Lambda$}
   \STATE Set pytorch seed to $s + r - 2$
   \STATE Standardize $D_{\text{train}},D_{\text{val}},D_{\text{test}}$ with sample mean and standard deviation of $D_{\text{train}}$
   \STATE Train $M$ on $D_{\text{train}}$ with hyperparameter setting $\lambda$, add the best validation hybrid loss $l_h(\cdot,D_{\text{val}};\alpha)$ achieved during training to the score of $\lambda$
   \ENDFOR
   \ENDIF
   \IF{$j=M$}
   \STATE Select $\lambda^*$ from $\Lambda$ with the lowest score
   \STATE Set pytorch seed to $s + r - 2$
   \STATE Standardize $D_{\text{train}},D_{\text{val}},D_{\text{test}}$ with sample mean and standard deviation of $D_{\text{train}}$
   \STATE Train $M$ on $D_{\text{train}}$ with hyperparameter setting $\lambda^*$, select the epoch $M^*$ with best validation hybrid loss $l_h(\cdot,D_{\text{val}};\alpha)$.
   \STATE Add $l_h(M^*,D_{\text{test}};\alpha=0)$ to $e_{\text{pred}}$
   \STATE Add $l_h(M^*,D_{\text{test}};\alpha=1)$ to $e_{\text{causal}}$
   \ENDIF
   \ENDFOR
   \ENDFOR
   \ENDFOR
   \STATE {\bfseries Output:} $e_{\text{pred}},e_{\text{causal}}$
\end{algorithmic}
\end{algorithm}
\section{Additional Experiments}
In this section we provide details of additional experiments that may be of interest to readers. We carried out three sets of additional experiments: (1) we apply novel interventions to the test set that are unseen in the training set; (2) we corrupt a small portion of training ranking information; and (3) we test hybrid models that are based on full (un-reduced) mechanistic models.  %\delrj{In these experiments, we only use one repeat for each value of $\alpha < 1$, to simplify computation; for $\alpha=1$, where the training procedure focuses solely on ranking (causal loss) and ignores predictive performance---leading to more variability in trained models based on initialization---we use the full 3 repeats.}

\subsection{Different Intervention Sets in Training and Test Sets}
\label{app:insulincarb}

\begin{figure}[h]
    \centering
    \includegraphics[height=3.5in]
    {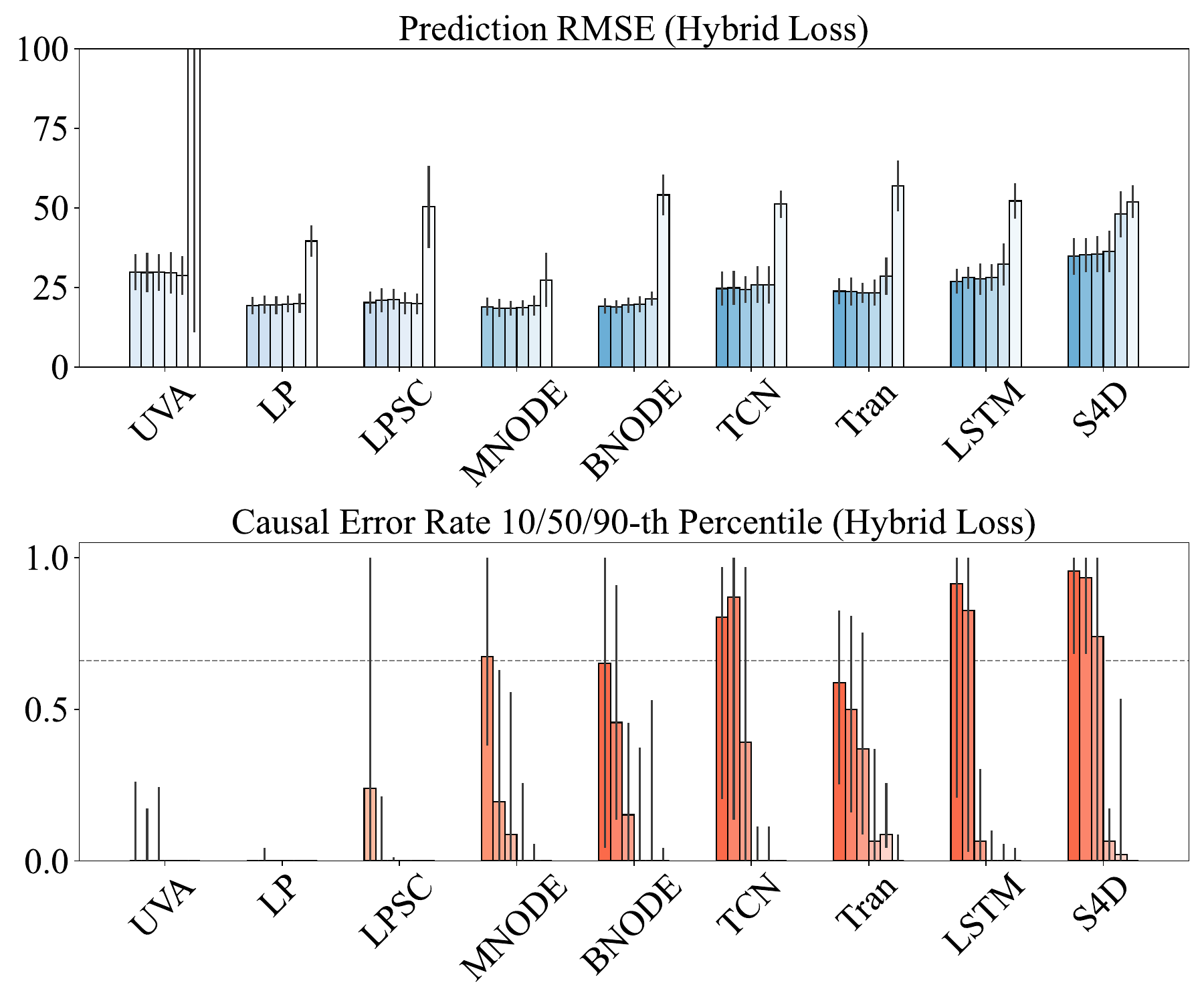}
    %{sections/img/insulincarb_ratio.png}
    \caption{As in Fig.~\ref{fig:synthetic}, but for T1DEXI data and using a full UVA/Padova mechanistic model baseline and hybridizations of the reduced UVA/Padova simulator for LP, LPSC, and MNODE.  In these experiments, the training data includes only interventions on either insulin or carb intake, but not both together; however, the test data includes interventions on the insulin-carb ratio.}
    \label{fig:inscarb_ratio}
\end{figure}

In this experiment, the training intervention sets consist of only category (1) and (2) described in Section \ref{sec:data} (Intervention Sets).  Crucially, note that our training intervention sets {\em do not} include interventions where both insulin and carbohydrates can vary, as in category (3) in Section \ref{sec:data}; this ensures that our training is {\em only} teaching the models about dependence on insulin alone and carbs alone.  On the other hand, the test intervention set contains only the following category: add 45 grams of carbohydrate at the end of exercise, and at the same time add 2.25/3.00/4.50 units of insulin.    These interventions correspond to a insulin-carb ratio of 1:20/1:15/1:10 respectively. We keep other experimental settings the same as the main T1DEXI experiments.
\paragraph{Results}
The results of the experiments are summarized in Figure \ref{fig:inscarb_ratio}.  The results illustrate that for moderate values of $\alpha$ that balance the prediction loss with the causal loss, the causal error rate of most models improves even in this setting where the test intervention set contains interventions not seen in the training data.  Further, we see that in general, the hybrid models offer more rapid improvement in causal loss as $\alpha$ increases, with virtually no increase in predictive loss.  BNODE appears to offer comparable performance to MNODE, but with generally higher variance in its causal loss.
%The results illustrate that for moderate values of $\alpha$ that balance the prediction loss with the causal loss, the hybrid models are able to generalize well even in this setting where the test intervention set contains interventions not seen in the training data.  The results suggest that the inductive bias provided by hybrid loss combined with hybrid modeling is allowing the joint relationship to of carbs and insulin to be effectively learned.  Note this is not the case for either the fully mechanistic model (which has high predictive loss) or most of the black box models (that exhibit high causal loss). \rj{I think we should consider rewriting this to emphasize hybrid loss more than hybrid models}

\subsection{Corrupted Ranking Knowledge in Training Set}
\label{app:corrupted}

\begin{figure}[h]
    \centering
    \includegraphics[height=3in]
{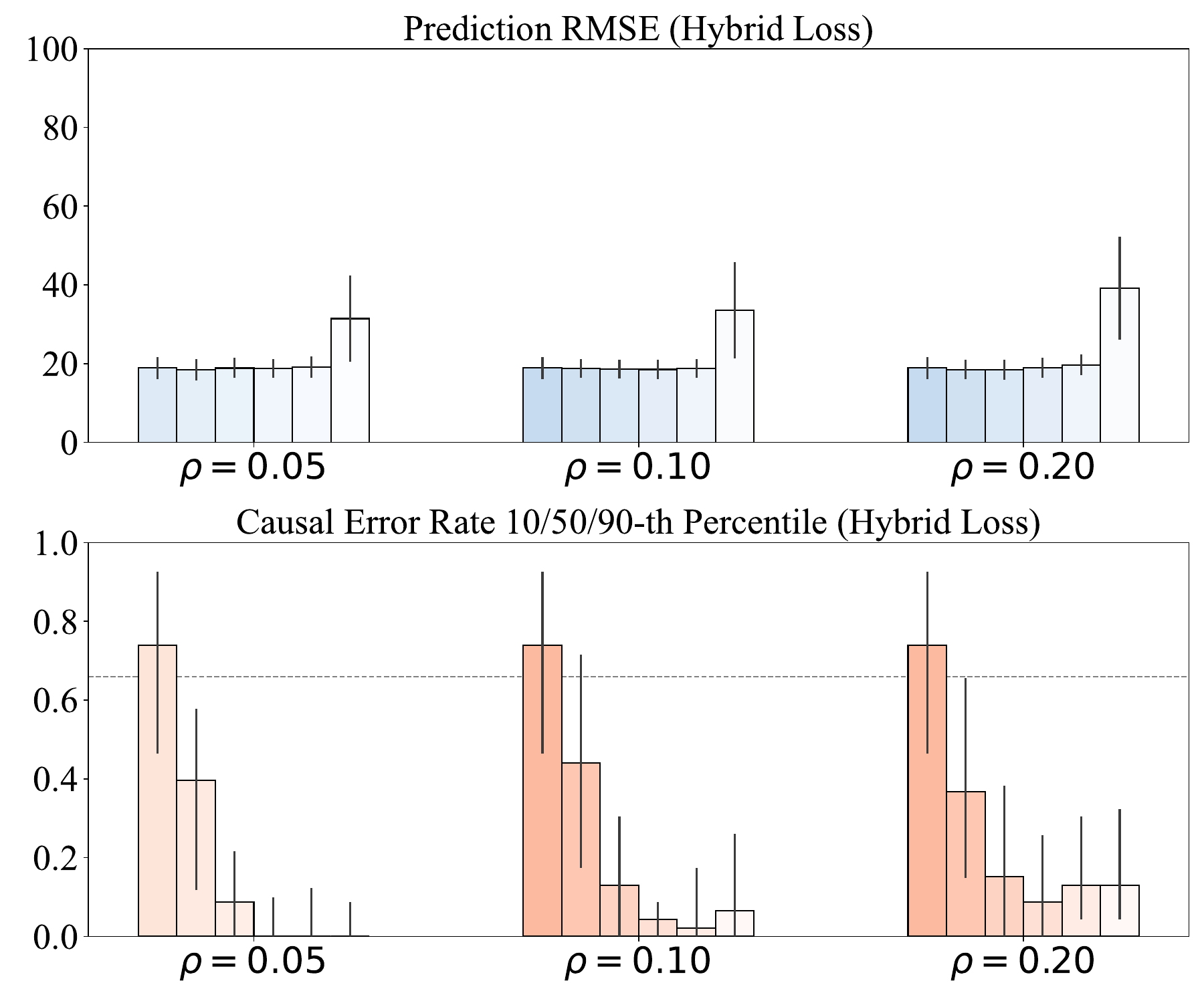}
    %{icml_camera_ready/sections/img/corruption.png}
    \caption{As in Fig.~\ref{fig:synthetic}, but for T1DEXI data with MNODE, using a hybridization of the reduced UVA/Padova simulator.  In these experiments, a fraction $\rho$ of the training data is corrupted.}
    \label{fig:corrupt}
\end{figure}

In this experiment, the ground truth ranking of each training intervention set in $\mathcal{I}^*$ has a probability of $\rho\in\{0.05,0.10,0.20\}$ of being corrupted by circularly shifting the ranking to the right by 1 position.  We carry out these simulations only for MNODE as
a representative example.  We keep the testing ground truth intact and other experimental settings the same as the main T1DEXI experiments.
\paragraph{Results}
The results of the experiments are summarized in Figure \ref{fig:corrupt}.  We see that at moderate values of $\alpha$, the performance on causal tasks degrades %\delebf{at moderate values of $\alpha$ as corruption increases} 
with increasing levels of corruption.  Notably, we also see that when the corruption rate is high, %\delebf{see that} 
{\em high} $\alpha$ (high weight on the causal loss) can actually {\em deteriorate} causal performance, 
%\delebf{lead to {\em worse} performance when corruption is high}, 
suggesting that the model is erroneously overemphasizing rankings from the corrupted data (introducing bias).  %\delebf{However} 
Regardless, even when the corruption rate is high,
choosing any $\alpha > 0$ (i.e., using a hybrid loss) lowers the causal error rate relative to a pure predictive loss ($\alpha = 0$).
%utilizing the hybrid loss is beneficial over the pure predictive loss ($\alpha=0$) regardless of the setting of $\alpha$.}

\subsection{Hybrid Models Based on Full UVA/Padova}
\label{app:hybrid_full}
In the main paper, we work with a reduced UVA/Padova model.  For completeness, we also implemented the latent parameter (LP) model and mechanistic neural ODE (MNODE) model using the full UVA/Padova model described in Appendix \ref{uva-appendix}.  Their implementations are exactly the same as described in Appendix \ref{implementation-detail} except that they use model equations/adjacency matrices of the full UVA/Padova model, and their hyperparameters are tuned on slightly different grids to count for the drastic increase in the number of mechanistic states and parameters. The search grid for hyperparameters of each full hybrid model is given below:
\begin{enumerate}
    \item LP: \[n=\{2,3\}\times m=\{16,24,32\}\times d=\{20,24,28\}\]
    \item MNODE: \[n=\{2,3\}\times m=\{16,32\}.\]
\end{enumerate}
Note that we did not include latent parameter learning with state closure (LPSC) in this set of simulations, as its full version requires an unreasonably large number of both states and model parameters when applied to the full UVA/Padova model.

\begin{figure}[h]
    \centering
    \includegraphics[height=3.5in]
    {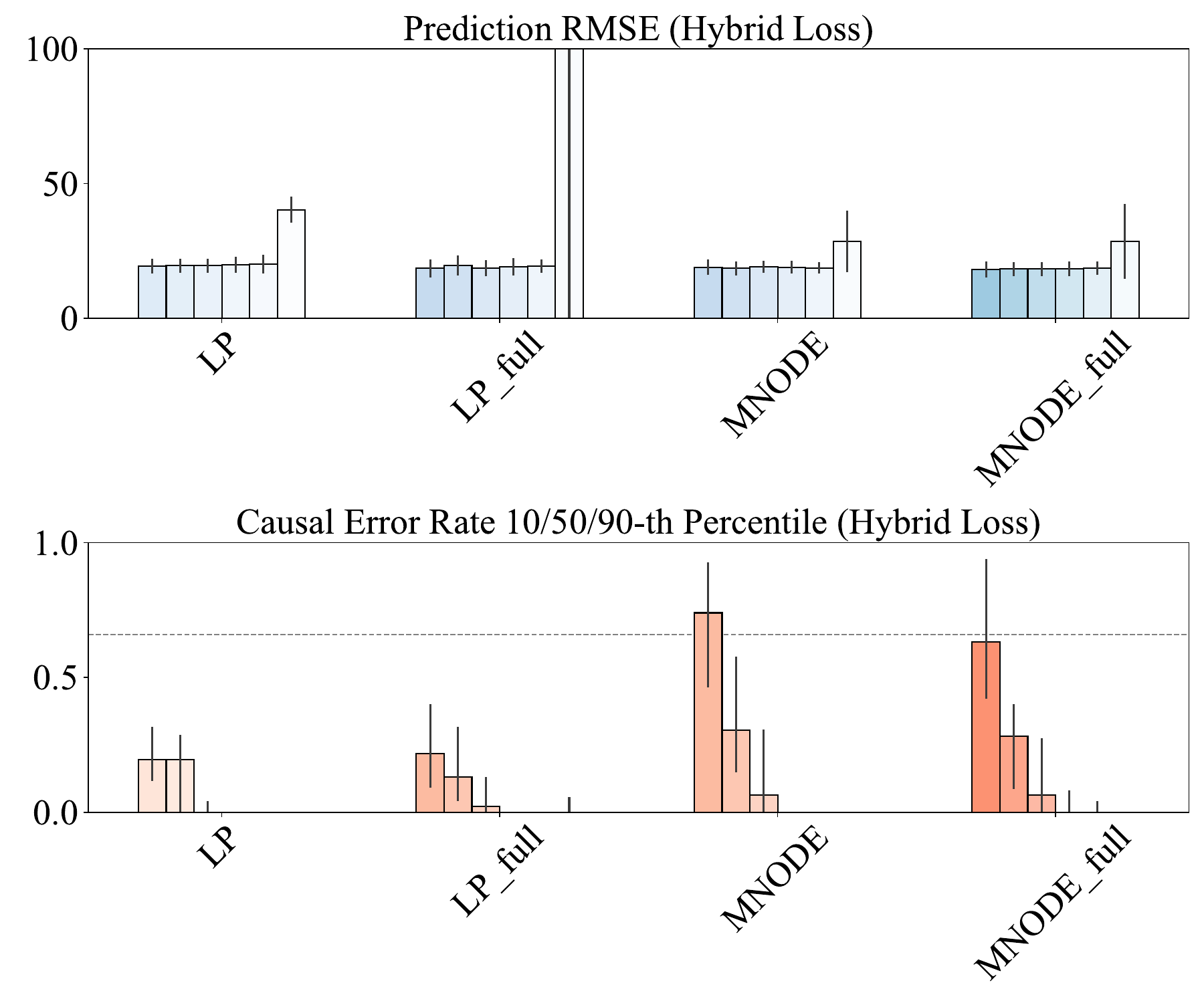}
    %{sections/img/full_experiments.png}
    \caption{As in Fig.~\ref{fig:synthetic}, but for T1DEXI data and comparing LP and MNODE hybrid models based on both the full and reduced UVA/Padova simulator as the mechanistic model.}
    \label{fig:full_exp}
\end{figure}

\paragraph{Results}
The results of the experiments are summarized in Figure \ref{fig:full_exp}.  We see that in general, the hybrid models based on the reduced UVA/Padova simulator offer  generalization performance on par with those of hybrid models based on the full simulator.  One of the primary benefits of building off of a reduced mechanistic model is that the resulting hybrid model has significantly reduced training time.  These results suggest model reduction is a valuable (but not necessary) implementation step in applying \modelname in practice.

\end{document}